\documentclass[9.4,conference]{IEEEtran}
\usepackage{booktabs,siunitx}
\usepackage{graphicx}
\usepackage{siunitx}
\usepackage{textpos}
\usepackage{bbm}
\usepackage{url}
\usepackage{multirow}
\usepackage{csquotes}
\usepackage{balance}
\usepackage{amssymb}  
\usepackage[numbers]{natbib}
\usepackage{mathtools}
\usepackage[numbered]{algorithm} 
\usepackage{algorithmicx}  
\usepackage{subcaption}
\usepackage{color}
 \usepackage{enumitem}
\usepackage[noend]{algpseudocode}
\floatname{algorithm}{Algorithm}

\newtheorem{definition}{Definition}[section]
\newtheorem{example}{Example}

\newcommand{\chainlet}[2]{\mathbb{C}_{#1 \rightarrow #2}}

\title{ChainNet: Learning on Blockchain Graphs with Topological Features}
\author{Nazmiye Ceren Abay, Cuneyt Gurcan Akcora, Yulia R. Gel, Umar D. Islambekov, Murat Kantarcioglu,  \\ Yahui Tian, Bhavani Thuraisingham

}
\begin{document}

\maketitle

\begin{textblock*}{500mm}(.15\textwidth,-6cm)
To Appear in the 2019 IEEE International Conference on Data Mining (ICDM).
\end{textblock*}

\begin{abstract}
With emergence of blockchain technologies and the associated cryptocurrencies, such as Bitcoin, understanding network dynamics behind Blockchain graphs has become a rapidly evolving research direction. Unlike other financial networks, such as stock and currency trading, blockchain based cryptocurrencies have the entire transaction graph accessible to the public (i.e., all transactions can be downloaded and analyzed). A natural question is then to ask whether the dynamics of the transaction graph impacts the price of the underlying cryptocurrency. We show that standard graph features such as degree distribution of the transaction graph may not be sufficient to capture network dynamics and its potential impact on fluctuations of Bitcoin price. In contrast, the new graph associated topological features computed using the tools of persistent homology, are found to exhibit a high utility for predicting Bitcoin price dynamics.  
Using the proposed persistent homology-based techniques, we offer a new elegant, easily extendable and computationally light approach for graph representation learning on Blockchain. 
 
\end{abstract}

\section{Introduction}
\label{sec:intro}

Recent jumps of Bitcoin price have led to ever growing debates with respect to the future of Bitcoin and cryptocurrencies and its potential impact on global financial markets~\cite{mattila2016blockchain}. One interesting aspect of popular cryptocurrencies such as Bitcoin is that each transaction is recorded on a distributed public ledger called blockchain. The recorded transactions can be then accessed and analyzed by anyone. Furthermore, all of the transactions could be represented by a graph referred to as the \textquote{blockchain graph}. Existence of the blockchain graph raises important questions such as \textquote{How does the blockchain graph structure impact the underlying cryptocurrency price?} 

In this paper, we focus on addressing this question by proposing different approaches to represent blockchain graph patterns; and we use these patterns to build machine learning models for Bitcoin price prediction. 

First approach that comes to mind to leverage the blockchain graph structure is to extract traditional graph features such as degree distribution, motif counts and clustering coefficients, and to use these graph features in machine learning models such as random forest for assessment of their utility in price forecasting. 
 
As already observed by previous studies (e.g., \cite{swanson2014learning, greaves2015using}), and also confirmed by our experimental results, these standard graph based features fail to capture important properties such as transaction volumes, transaction amounts, and their relationships with the underlying graph structure. Since these basic approaches do not provide conclusive insights into the blockchain graph dynamics and its impact on cryptocurrency price, we propose novel techniques inspired by topological data analysis (TDA) and, particularly, persistent homology that can capture these higher order interactions.  

\textit{Persistent homology} allows us to extract topological information from a blockchain graph and unveil some critical characteristics behind its functionality. Most notably, persistent homology captures interactions of the graph components at a multi-scale level which are otherwise largely inaccessible with conventional analytic methods. Such an approach provides the following important benefits. First, we systematically account for changes in the blockchain graph topology and geometry at different scales, both in terms of transaction patterns and associated transaction volumes. Second, by computing topological features for a range of scale values we bypass the problem of optimal scale selection. That is, instead we systematically derive topological information from the blockchain graph and use its change dynamics for cryptocurrency price prediction.
 
Third, the multi-scale approach permits us to effectively distinguish true topological features from noisy ones in a robust way based on the extent of feature lifespan across scale values. Furthermore, a few studies on the application of TDA to other types of networks show that persistent homology-based features outperform conventional graph features such as betweenness centrality, clustering coefficient and nodal degree in network classification and segmentation~\cite{AmanmeetPD}.

Our contributions can be summarized as follows:
\begin{itemize}[leftmargin=0.1cm,itemindent=.5cm,labelwidth=\itemindent,labelsep=0cm,align=left] 
\item To our knowledge, we are the first ones to introduce persistent homology to 
cryptocurrency predictive analytics. Furthermore, we couple homology-based topological features of Blockchain with machine learning techniques to predict Bitcoin prices.

\item We introduce a novel concept of a {\it Betti derivative}. 
Betti derivatives capture the rate of changes that occur in the topological structure of the blockchain graph. We show predictive utility of the Betti derivatives in forecasting Bitcoin prices. 
 
\item Using extensive empirical analysis, we show that machine learning models incorporating our proposed persistent homology-based methodology can significantly outperform (i.e., up to 38\% improvement in root mean squared error) models which use only past price and standard features such as total transaction count.
\end{itemize}
The remainder of the paper is organized as follows: In Section~\ref{sec:related}, we discuss the related work and emphasize the differences of our proposed approach. We discuss the background information related to blockchain graph representations in Section~\ref{sec:fl}, and persistent homology in Section~\ref{sec:betti}. In Section~\ref{sec:exp}, we present the experimental results. Finally, in Section~\ref{sec:concl}, we conclude by discussing the implications of our results with respect to cryptocurrency price dynamics and underlying blockchain graph structure.

\section{Related Work} 
\label{sec:related}
The success of Bitcoin~\cite{nakamoto2008bitcoin} has encouraged hundreds of similar digital coins~\cite{tschorsch2016bitcoin}. The underlying Blockchain technology has been adopted in many use cases and applications.  With this rapidly increasing activity, there have been numerous studies analyzing the blockchain technology from different perspectives.   

The earliest results aimed at tracking the transaction network to locate coins used in illegal activities, such as money laundering and blackmailing \cite{androulaki2013evaluating,ober2013structure}. These findings are known as the taint analysis~\cite{di2015bitconeview}.  

The Bitcoin network itself has also been studied from multiple aspects. Dyhrberg \cite{dyhrberg2016bitcoin} studied Bitcoin's similarities to gold and the dollar, finding hedging capabilities and advantages as a medium of exchange. From a graph perspective, Baumann et al.~\cite{baumann2014exploring} analyzed centralities, and \cite{lischke2016analyzing} found that since 2010 the Bitcoin network can be considered a scale-free network. Furthermore, \cite{kondor2014rich} tracked the evolution of the Bitcoin transaction network, and modeled degree distributions with power laws. Although these studies analyzed the Bitcoin graph, the primary focus was on global graph characteristics.  

Kristoufek~\cite{kristoufek2015main} analyzed  potential drivers of Bitcoin prices, such as the impact of speculative and technical sources. A number of recent studies show the utility of global graph features to predict the price \cite{kondor2014inferring,greaves2015using,madan2015automated}. For instance, \cite{sorgente2014reaction} studied the impact of average balance, clustering coefficient, and number of new edges on the Bitcoin price. These findings suggest that certain network features are correlated with price; for example, the number of transactions put into a block indicates a price increase. 

Community detection on weighted networks~\cite{allertonJogL15} has not been applied to blockchains yet, but two network flow measures were recently proposed by \cite{yang2015bitcoin} to quantify the dynamics of the Bitcoin transaction network and to assess
the relationship between flow complexity and Bitcoin market variables. Furthermore,~\cite{madan2015automated} identified 16 features (e.g., number of Tx) for 30, 60 or 120 minute intervals and used random forest models to predict the price. The core idea behind all these approaches is to extract certain global network features and to employ them for predictions. However interactions of features~\cite{henelius2016finding} are not widely studied.  
Most recently,~\cite{akcora2018chainlet} introduced the notion of \textit{chainlet} motifs to understand the impact of local topological structures on Bitcoin price dynamics, and showed that
employing aggregated chainlet information leads to more competitive price prediction mechanisms. In contrast to global network features, chainlets provide a finer grained insight at the network transactions. However,
the chainlet approach of~\cite{akcora2018chainlet} is limited to analysis of transaction types and does not account for critical information such as the transferred amounts. In this paper, we remedy some of these short comings using persistent homology based features which 
yields more competitive performance, with more than three times improvement over the highest gain reported in~\cite{akcora2018chainlet}.

\section{Learning Graph Based and Topological Features}
\label{sec:learning}
{\noindent
{\begin{minipage}{0.98\linewidth}
\textbf{Problem Statement:} Let $x_{t}\in \mathbb{R}^d$ be a set of features computed on the Bitcoin blockchain. Let $(x_1,y_1),\ldots,(x_t,y_t)$ be the observed data where $Y=\{y_1,\ldots,y_t\}$ are the corresponding Bitcoin prices in dollars. At a time point $t$, estimate the Bitcoin price $y_{t^\prime}$ where $t^\prime>t$.
\end{minipage}} 
}\newline
 
 \noindent To address this problem, we need to answer the following questions: 
\begin{itemize}[leftmargin=0.1cm,itemindent=.5cm,labelwidth=\itemindent,labelsep=0cm,align=left] 
    \item {\it How can real world Bitcoin prices be determined by blockchain network activity? Can the causality be proven?}
    \item[] $\blacklozenge$ Our hypothesis is that input and output based structure of Bitcoin transactions encode various buyer and seller motivations that reflect market sentiment, which in turn determines price movements. For example, investments in the currency are encoded in transactions that contain more inputs than outputs. Similarly, selling behaviour creates transactions with more outputs than input addresses. Already, previous results~\cite{akcora2018chainlet}  offer evidence for a causality between blockchain activity and Bitcoin price. In this work, we offer further evidence for the causality.
     
    \medskip 
    \item {\it Most Bitcoin transactions on online exchanges are handled in-house by exchanging private/public keys pairs between users. How can we account for these missing transactions?}
    \item[] $\blacklozenge$ We are aware that in-house transactions can be as many as 3 to 30 times (See Figure 4 in ~\cite{antulov2018inferring}) the number of transactions published in the blockchain. However, we claim that in-house transactions are still periodically published to the blockchain in batches. Transaction histories of exchange addresses, such as the Coinbase Bitcoin address,~\footnote{{\small\url{https://www.blockchain.com/btc/address/1LQTXi1iWULMd4aKn5tKpcgT3xgJiTV5Dm}}} contain evidence to support our claim.  Otherwise a data loss would bring about huge losses, as happened to the Mt. Gox exchange in 2014. Although they contain a lagged version of data, exchange transactions still contain useful information, and their amounts can be utilized in a predictive model.
  
  \medskip 
  \item {\it From a methodological perspective, why is the price prediction problem important?}  
  \item[] $\blacklozenge$ Price prediction is important as price dynamics impacts a billion dollar industry in cryptocurrencies. 
   Furthermore, we argue that price, which is arbitrated off-chain in real world, is a unique external validator for testing the power of machine learning models on a complex system that is created worldwide by real actors. For example, in this work we use the price to validate the predictive power of TDA tools and summaries, e.g., \textit{Betti derivatives}. As price is inherently related to real life phenomena, such as network growth and influential user behaviour, we envision that many network growth, scaling  and influence models~\cite{gionis2012estimating} can be validated by using settings similar to ours.  
\end{itemize}

\noindent We provide two solutions to our research problem: {\it graph filtration~(FL)} and the {\it Betti sequences}. 
The first approach is based on graph filtration. That is, we filter the transaction network with increasing thresholds of Bitcoin amounts, and create multiple realizations of the network. Afterwards, we merge these realizations to train a model. The second approach uses topological summaries to capture persistent features in terms of Betti sequences and Betti derivatives. 

The Betti approach is based on rigorous mathematical foundations of algebraic topology and provides a multi-lens view of the system, whereas the graph filtration is a heuristic that allows manually selecting amount thresholds and associated filtering of the network.  Next, we describe these two approaches in details.

\subsection{Learning Graph Representations}
\label{sec:fl}
We first introduce existing blockchain network models and explain their shortcomings. Next we describe our substructure model of the blockchain graph and extract \textit{graph filtration} features. 

In a typical blockchain graph such as the one used by Bitcoin, an owner of multiple addresses (i.e., each address represents an account, each person may have many addresses/accounts) can combine them in a transaction and send coins to multiple output addresses. Therefore, the Bitcoin blockchain consists of two types of nodes: transactions, and addresses that are input/output of transactions.
Earlier results on Blockchain analysis are based on constructing graphs with a single type of node:  \textit{transactions}~\cite{ron2013quantitative} or \textit{addresses}~\cite{filtzevolution} constituted nodes and currency transfers created edges between nodes. By choosing a single type of node, these approaches omit either address or transaction information in the graph. In our approach we follow~\cite{akcora2018chainlet} and
construct a heterogeneous Blockchain graph with both  address and transaction nodes.  Note that the Blockchain edges are naturally ordered in time with respect to the block they appear in. 
Once the graph is constructed, shapes of transactions, and how they connect addresses conveys information on how the graph further extends in time. For all purposes, a Blockchain graph can be thought as a forever forward branching forest where transaction nodes appear only once, and address nodes may appear multiple times (but in practice address reuse is discouraged on Bitcoin). 

With its input and output addresses, each transaction represents an immutable decision that is encoded as a substructure on the blockchain graph. Recently, \cite{akcora2018chainlet} proposed to study such blockchain substructures in the form of \textbf{chainlets}.

\begin{definition}[The $k$-Chainlet~\cite{akcora2018chainlet}]
Let $\mathcal{G} = (V, E, B)$, be the directed, heterogeneous blockchain graph, where
$V$ is a set of vertices, $E \subseteq V\times V$ is a set of directed edges, and $B=$\{\textbf{Address}, \textbf{Transaction}\} represents node types. A blockchain subgraph $\mathcal{G'}=(V', E',B)$ is a \textit{subgraph} of $\mathcal{G}$ (i.e., $\mathcal{G'}\subseteq \mathcal{G}$), if $V'\subseteq V$ and $E'\subseteq E$.  Let $\mathcal{G}_k=(V_k, G_k,B)$ be a subgraph of $\mathcal{G}$ with $k$ nodes of type $\left\{\textbf{Transaction} \right\}$. The $\mathcal{G}_k$ is called a graph \textit{$k$-chainlet}. 
For a graph chainlet if there exists a $G_k \in G$, we say that there exists an \textbf{occurrence}, or \textit{embedding} of $\mathcal{G}_k$ in $\mathcal{G}$.
\end{definition}

The chainlet approach of~\cite{akcora2018chainlet}
aims to transfer the ideas of network motifs~\cite{milo2002network} to blockchain graphs. That is, by counting frequency of certain shapes, a blockchain graph can be summarized with chainlet densities. However, while the chainlet approach of~\cite{akcora2018chainlet} is found to be promising in describing dynamics of the blockchain graph, it has two major shortcomings. First, \cite{akcora2018chainlet} focuses only on the basic case of $k=1$, or 1-chainlets. Indeed, as
the $k$ value increases, $k$-chainlets encode higher order structures on the graph and the number of distinct shaped chainlets also increases. As each transaction can have thousands of inputs and outputs, even for the most basic case of $k=1$, $k$-chainlets can have millions of distinct shapes. Second, even in the basic case of 1-chainlets, \cite{akcora2018chainlet} disregards such critical information as amounts of coins transferred from its inputs to outputs. In this paper, we address the second shortcoming and incorporate the key information on the transferred amounts into analysis of blockchain substructures.

\paragraph{Occurrence and Amount Matrices} 
 On the Bitcoin network, the output and input addresses of a transaction $t_n$ are defined as a list of addresses $\left |\Gamma^{o}_n\right| \geq 1$ and $\left |\Gamma^{i}_n\right| \geq 1$, respectively. An address $i_a \in \Gamma^{i}_n$ has an associated coin amount $A({i_a})$ that $t_n$ receives. The output amount of a transaction $t_n$ is defined as the sum of outputs from all input addresses $\mathcal{A}^{o}(n)=\sum_{i_a \in \Gamma^{i}_n}{A(i_a)}$.
Considering all transactions $T$, we define the maximum number of inputs, $i_{max}=\underset{t_n \in T}{argmax}{(\mathrm{\left | \Gamma^i_n\right|})}$ and outputs $o_{max}=\underset{t_n \in T}{argmax}{(\mathrm{\left | \Gamma^o_n\right|})}$.
 
 We then encode chainlet substructures with two dimensions: for $\left |i\right|$ \textit{input} addresses and $\left |o\right|$ \textit{output} addresses, the chainlet is denoted as $\chainlet{i}{o}$. The blockchain graph can be then represented in a form of two matrices, that is, the occurrence
$\mathcal{O}_{[i_{max}\times o_{max}]}$ and amount $\mathcal{A}_{[i_{max}\times o_{max}]}$ matrices, where the cell of $i$-th row and $o$-th column represents information on the substructure $\chainlet{i}{o}$. 

\begin{figure}
\centering
  \includegraphics[width=0.7\linewidth]{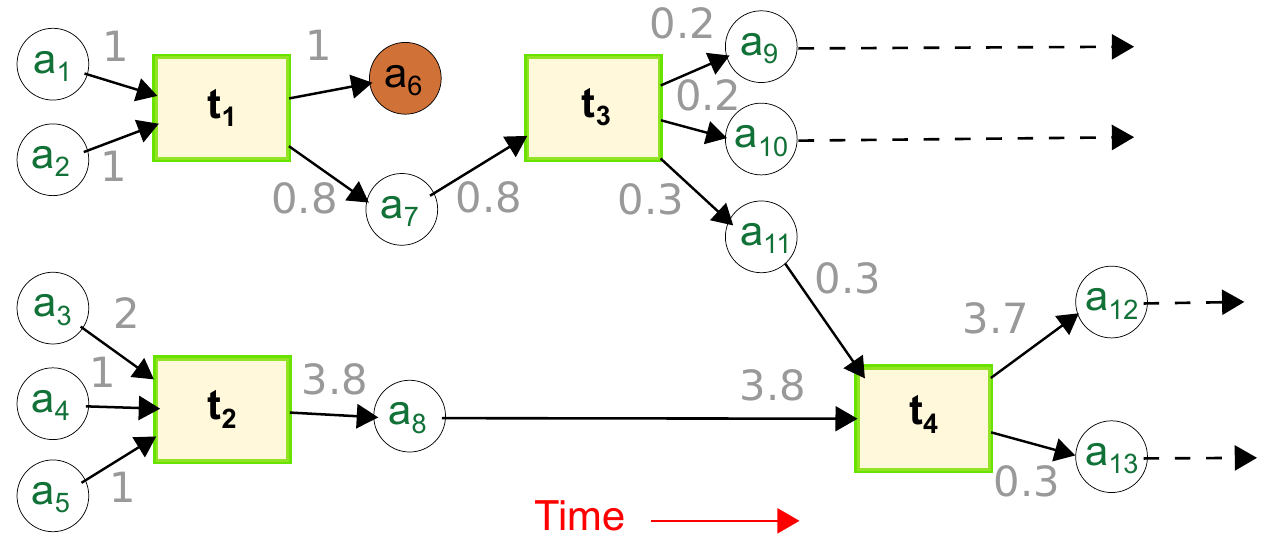}\caption{A Bitcoin graph with 4 transactions and 13 addresses. Amounts on edges show currency transfers. The difference between input and outputs amounts, if exists, shows the transaction fee collected by miners.}
\label{fig:1blockchaingraph}
\end{figure}

\begin{example}
Consider the toy example in Figure~\ref{fig:1blockchaingraph}, 
where both $i_{max}=3$ and $o_{max}=3$. Resulting $3\times3$ occurrence and amount matrices are given below as $\mathcal{O}$ and $\mathcal{A}$, respectively. In total, there are four chainlets but only three distinct shapes. $\chainlet{1}{3}$ and $\chainlet{3}{1}$  occurs once ($\mathcal{O}_{13}=\mathcal{O}_{31}=1$), and $\chainlet{2}{2}$ occurs twice ($\mathcal{O}_{22}=2$). The total amounts transferred by each chainlet are given as $\mathcal{A}_{13}=0.8$, $\mathcal{A}_{22}=4.1+2$ and $\mathcal{A}_{31}=3.8$. 
\[
\mathcal{O}=\begin{bmatrix}
    0       & 0 & 1  \\
    0       & 2 & 0  \\
    1       & 0 & 0
\end{bmatrix}
~\text{and}~
\mathcal{A}=\begin{bmatrix}
   0 & 0 & 0.8 \\
   0 & 6.1 & 0 \\
   4 & 0 & 0
\end{bmatrix}
\]
\end{example}

\paragraph{Graph Filtration (FL)}
Given the amount and occurrence information, a natural combination of them entails filtering the occurrence matrix with user defined thresholds on amounts, or filtering the amount matrix with user defined thresholds on occurrences. In both cases, the user defined threshold implies a heuristic aspect.

\begin{algorithm}
\footnotesize
    \caption{FL: Graph Filtration}
    \label{alg:FL}
    \begin{algorithmic} [1]
        \Require $\mathcal{G}$: Blockchain graph, time $t$, $\epsilon_{1,..S}$: set of $S$ filtration scales.
        \For {$\epsilon \in \epsilon_{1,..S}$}
        \State $\mathcal{O}^\epsilon \leftarrow \left[ \right]$ //initialize occurrence matrix
        \EndFor
       
        \For {chainlet $\chainlet{i}{j} \in \mathcal{G}_t$ } 
        \For {each scale $\epsilon \in \epsilon_{1,\dots,S}$}
            		\If {$\epsilon \leq amount(\chainlet{i}{j})$} 	
                		\State $\mathcal{O}^{\epsilon}_{ij} \leftarrow 1+\mathcal{O}^{\epsilon}_{ij}$                 	\EndIf     
                
            \EndFor
            
    \EndFor
    
    \State \Return{$x_t= \left [ \mathcal{O}^{\epsilon_1}   \ldots  \mathcal{O}^{\epsilon_S} \right ] $// concatenated occ. matrices }
    \end{algorithmic}  
\end{algorithm}

FL creates multiple occurrence matrices of a Bitcoin network at a given time period, and uses them as the feature set to train a prediction model. Algorithm~\ref{alg:FL} represents the main steps. At a given time period $t$, chainlets of the time period are iterated over with a set of thresholds. A chainlet $\chainlet{i}{j}$'s occurrence is recorded in the associated occurrence matrix $\mathcal{O}^{\epsilon}$ if the amount transferred by the chainlet $amount(\chainlet{i}{j}) \geq \epsilon$. The process is repeated for all inputted data. Resulting occurrence matrices are row-wise concatenated and output as the FL feature set for time period $t$ (i.e., $x_t$). 

The FL captures persistent graph substructures by retaining edges among nodes according to a set of scale values. For a scale value $\epsilon \in \epsilon_{1,\ldots,S}$, we only record the occurrence of chainlet substructures, if the amount transferred by the substructure is  $\geq \epsilon$.

\subsection{Learning Topological Representations} 
\label{sec:betti}

We start from summarizing the conventional TDA tools and then proceed to the proposed
TDA-based methodology for blockchain graph
analytics.

TDA is an emerging field at the intersection of algebraic topology and computational geometry providing methods to systematically study the topological and geometric structure underlying data \cite{Carlsson,TDAintro}. In this context, these structures are commonly analyzed via the multi-scale-based framework of persistent homology. Below we outline its main steps. The primary idea is to assess which topological features remain persistent over a larger set of scales and hence, e.g., in the case of the Blockchain network, are likely to play a significant role in its functionality.

Let $\mathbb{X}=\{X_1, \ldots, X_n\}$ be a set of data points in a metric space (e.g., the Euclidean space).
Select a scale $\epsilon_k$ and form a graph $G_k$ with the associated adjacency matrix $A=\mathbbm{1}_{d_{ij} \leq \epsilon_k}$, where $d_{ij}$ is the distance between points $X_i$ and $X_j$. Changing the scale values $\epsilon_1<\epsilon_2<\ldots<\epsilon_N$ results in a hierarchical nested sequence of graphs $G_1 \subseteq G_2 \subseteq \ldots \subseteq G_N$ that is called a \textit{graph filtration}. 

Next, to be able to glean the intrinsic geometry underlying the data from the graph filtration, we associate an \textit{(abstract) simplicial complex} with each $G_k$, $k=1,\ldots, N$. These constructs can be thought of as higher order analogues of graphs having both the topological and combinatorial structure \cite{TDAintro}. The latter serves well for the computational purposes to extract various topological summaries from data. A major advantage of the multi-lens perspective is that it avoids the issue of searching for an optimal scale value and associated feature engineering. 

The choice of a simplicial complex to be adopted depends on the complexity of the data and which topological features one is interested in highlighting. The \textit{Vietoris-Rips} (VR) simplicial complex is one of the most popular choices in TDA due to its easy construction and computational advantages \cite{Carlsson, Zomorodian:2010}. 

\begin{definition}[Vietoris-Rips complex]
A \textit{Vietoris-Rips complex} at scale $\epsilon$, denoted by $VR_\epsilon$, is the abstract simplicial complex consisting of all $k$-element subsets of $\mathbb{X}=\{X_1, \ldots, X_n\}$, called ($k-1$)-simplices, $k=1,\ldots, K$, whose points are pairwise within distance of $\epsilon$. A 0-simplex can be identified with a point, a 1-simplex with a segment, a 2-simplex with a triangle and a 3-simplex is with a tetrahedron and so on. 
\end{definition}

Armed with the associated VR filtration, $VR_1 \subseteq VR_2 \subseteq \ldots \subseteq VR_N$, we can track qualitative topological features such as connected components, loops and voids that appear and disappear as we move along the filtration.

In our analysis, we use the Betti sequences as summaries of persistent homology calculations which encode the counts of these features at increasing scale values. Their individual elements are called the \textit{Betti numbers} that are computed for each value of the scale:
$$
\boldsymbol\beta_p=(\beta_{p}(\epsilon_1),\beta_{p}(\epsilon_2),\ldots,\beta_{p}(\epsilon_N)), \ \ \ p=0,1,\ldots,K,
$$
where $\beta_p(\epsilon_k)$ is the $p$-th Betti number of the simplicial complex at scale $\epsilon_k$. The Betti numbers for small $p$ have a simple interpretation. For instance, $\beta_0$ is the number of connected components; $\beta_1$ is the number of loops; $\beta_2$ is the number of voids etc. Formally, the Betti numbers are defined as follows:  

\begin{definition}[Betti numbers]
The $p$-th Betti number $\beta_p$, $p\in Z^{+}$, of a simplicial complex is the rank of the associated $p$-th \textit{homology} group defined as the quotient group of the \textit{cycle} and \textit{boundary} groups. 
\end{definition}

\subsubsection{Betti Sequences for a Blockchain Network}
\label{sec:bettinumbers} 
Although the Betti sequences provide a non-parametric solution to combine information on edge distance with node connectedness, the computational complexity of Betti calculations prohibits their usage in large networks. For example, for simplicial complexes of dimension 2, \textquote{currently no upper bound better than a constant times $n^3$ is known}~\cite{edelsbrunner2014computational}. For Betti numbers $\beta_{p>3}$, the complexity becomes too restrictive. This problem is compounded in the Bitcoin network because address reuse is discouraged. As such, every day brings more than 500K new nodes to the network. Betti number computations on such large networks is unfeasible. 

To solve the complexity issues, we propose a novel approach that computes the Betti sequences on a network of $N\times N$ nodes where N is the size of the amount matrix $\mathcal{A}$ (See Section~\ref{sec:fl}). Each of the $N^2$ unique chainlets (e.g., $\chainlet{2}{3}$) creates a node in the new network, where edge distance between two nodes is computed with a suitable 'distance' $d$. We describe the main steps as follows: 

Given a heterogeneous Blockchain network with transferred bitcoins on edges, 
\begin{enumerate}
	\item All the transferred amounts are converted from Satoshis to bitcoins (dividing by $10^8$), then added one (so that the values after taking logarithm are non-negative) and log-transformed: $a^\prime=\log(1+a/10^8)$, where $a$ is an amount in Satoshis. 
\item For each chainlet of a given time period, we compute the sample $q$-quantiles for the associated log-transformed amounts \cite{hyndman1996sample}: a $k$-th $q$-quantile, $k=0,1,\ldots,q$, is the amount $Q(k)$ such that
$$\sum_{i=1}^\tau \mathbbm{1}_{y_{i} < Q(k)}\approx\frac{\tau k}{q} \hbox{  and  } \sum_{i=1}^\tau \mathbbm{1}_{y_{i} > Q(k)}\approx \frac{\tau(q-k)}{q},$$
where $\tau$ is the total number of transactions. The (dis)similarity metric $d_{ij}$ between chainlet nodes $i$ and $j$ is defined as
the quantile-based distance
    $$d_{ij}=\sqrt{\sum_{k=0}^{q}[Q_i(k)-Q_j(k)]^2}.$$
	\item We construct a sequence of scales $\epsilon_1<\epsilon_2<\ldots<\epsilon_S$ covering a range of distances during the entire 365-day period. For each $\epsilon_k$, we build the corresponding VR complex whose 0-simplices are single chainlets and 1-simplices are pairs of chainlets with distance $\leq \epsilon_k$. As a result, we obtain the filtration of VR complexes $VR_1 \subseteq VR_2 \subseteq \ldots \subseteq VR_{S}$.
\item Armed with the VR filtration, we then compute $x_t = \{\beta_0(\epsilon_1),\ldots, \beta_0(\epsilon_S);\beta_1(\epsilon_1),\ldots, \beta_1(\epsilon_S)\}$.

\end{enumerate}

In constructing the new network, we use and hence retain the amount information from the Blockchain network. Furthermore, each node type (chainlet substructure) encodes the number of inputs and outputs in a transaction. This way, we combine distance (computed from transferred coins) with edge connectedness while restricting the network size. Our new TDA approach can work with networks of any size, and our experimental results (See Section~\ref{sec:exp}) show predictive power of its topological features.

\subsubsection{Betti derivatives}
\label{sec:derivative}
The graph of the $p$-th Betti sequence is often referred to as the \emph{$p$-th Betti curve}. Analysis of the Betti curves allows us to assess dynamics of essential topological features as a function of the scale.  Furthermore, to assess the rate of changes in topological features of the Blockchain graph, we introduce a novel concept of \emph{Betti derivatives} up to order $\ell>0$ on VR filtrations:  
$$\Delta^{\ell}\beta_p(\epsilon_k)=\Delta^{\ell-1}\beta_p(\epsilon_{k+1})-\Delta^{\ell-1}\beta_p(\epsilon_{k}),$$
where $k=1,2,\ldots,S-1$, $p=\{0,1,\ldots\}$ values are determined by how many Betti numbers we choose to use, and $S$ is the number of filtration steps. These finite differences are analogues of derivatives for smooth functions. The inclusion of the rates of change of the Betti curves is intended to systematically capture dynamics of essential topological features and to enhance the predictive power. In \cite{Hofer2018} the topological features of dimension zero are split into the essential (persisting till the end of filtration) and non-essential. However, there could be features that persist over a significant range of scale values but disappear right before the filtration ends and thus fall under the category of non-essentials. In contrast, our approach considers the Betti curves along with their shape rate derivatives as a whole and thereby allows to view such features under a more general umbrella of the essential features. 

\section{Experiments}
\label{sec:exp}

In this section, we show the performance of predictive models in our ChainNet framework.  

\subsection{Data}
\label{sec:dataset}
 We downloaded and parsed the entire Bitcoin transaction graph from 2009 January to 2018 December. Using a time interval of 24 hours, we extracted daily transactions on the network and created the Bitcoin graph. Our Bitcoin price (USD) data is downloaded from \url{blockchain.com} which aggregates prices from worldwide online exchanges.~\footnote{Due to the extreme divergence in prices from the rest of the world, Korean exchanges are excluded in Bitcoin price arbitration.} 

\paragraph{Filtration data.}
We analyzed Bitcoin transactions to find an appropriate dimension $N$ for the occurrence matrix. On the Bitcoin graph \% 90.50 of the chainlets have $N$ of 5 (i.e., $\chainlet{i}{o}$ s.t., $i < 5$ and $o < 5$) in average for daily snapshots. This value reaches \% 97.57 for $N$ of 20. We chose $N=20$, because it can distinguish a sufficiently large number (i.e., 400) of chainlets, and still offer a dense matrix. 

Our models achieved a satisfactory performance with  $\epsilon \in \{0,10,20,30,40,50\}$ scales in the graph filtration. However we note that $\epsilon$ partitions can be further improved.

\paragraph{Betti and Betti Derivative Data.} 
We use the Betti numbers estimation routine of the Perseus \cite{Perseus} software which provides an efficient algorithm to compute the Betti numbers and persistent intervals. 

We used $S \in \{50, 100, 200\text{ and }400\}$ as the filtration length. Overall, we find no improvement in prediction accuracy for $S>400$. Furthermore, there is no single optimal value of $S$ to be used in all statistical and machine learning models.

\begin{figure}
\centering
  \includegraphics[width=0.6\linewidth]{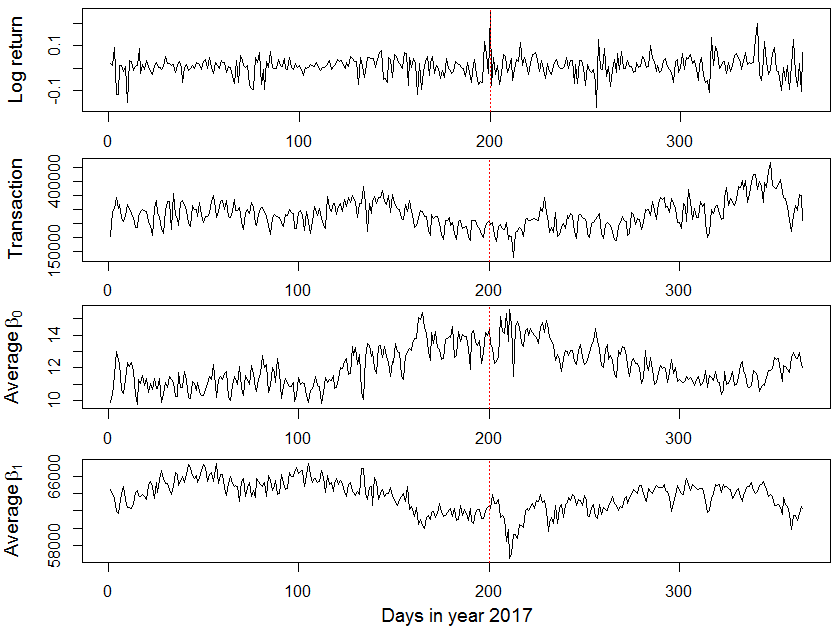}
  \caption{Time series of daily log returns, transactions, average $\beta_0$ and $\beta_1$ numbers in 2017.}
  \label{fig:Avrbetti}
\end{figure}

To decrease computational costs, in the present study, we focus on VR complexes of dimension one. This implies that the loops are formed by three or more nodes, which in turn leads to a general negative association between the Betti-0 and Betti-1 curves -- as $\epsilon$ increases, more simplices are added to the complex, thereby reducing the number of connected components and increasing the number of loops. For the same reason, we see in Figure \ref{fig:Avrbetti} that the spikes in average Betti-0 curves match the plummets of the corresponding Betti-1 curves and vice versa. On July 20, 2017 the Bitcoin Improvement Proposal 91, to trigger Segregated Witness (SegWit) activation, is locked in. This has resulted in the start of the new bullish wave. Remarkably, we find that the spike in Bitcoin in mid July 2017 have been preceded by an increase in Betti-0, and decreases in Betti-1 and average daily transactions. Moreover, the extrema of Betti-0, Betti-1 curves and average daily transactions in July 2017 are well aligned.

In addition to FL and Betti related features, we also experimented with basic features: price, mean degree of addresses (MeanDegree), number of new addresses (NumNewAddress), mean and total coin amount transferred in transactions (meanTxAmount and TotalTxAmount, respectively) and address network average clustering coefficient (ClusCoeff). Among these, we only found Price and TotalTx to be useful predictors and included them in our models.  Table~\ref{tab:dataExplanation} shows all the considered features. 

\begin{table}[htbp]
  \centering
  \caption{Features used in Machine Learning models for a given day.}
    \begin{tabular}{ll}
    \footnotesize
    \textbf{Approach} & \textbf{Feature Set} \\ \hline
    {Basic features } & $Price, TotalTx, MeanDegree$\\
    &$MeanTxAmount,TotalTxAmount$ \\ 
     & $NumNewAddress, ClusCoeff$\\
     \hline
    {Filtration (Sec~\ref{sec:fl})} & $Price, TotalTx, \mathcal{O}^{\epsilon_1}   \ldots  \mathcal{O}^{\epsilon_S}$\\ \hline
    {\multirow{1}[0]{*}{Betti (Sec.~\ref{sec:bettinumbers})}} & $Price, TotalTx, \beta_0(\epsilon_1),\ldots, \beta_0(\epsilon_S)$\\ & $\beta_1(\epsilon_1),\ldots, \beta_1(\epsilon_S)$\\
    \hline
    \multicolumn{1}{l}{Betti derivative}  & $ Price, TotalTx$ \\ & $\beta_0(\epsilon_1),\ldots, \beta_0(\epsilon_S),\beta_1(\epsilon_1),\ldots, \beta_1(\epsilon_S),$ \\ (Sec.~\ref{sec:derivative}) &$\beta^\prime_0(\epsilon_1),\ldots, \beta^\prime_0(\epsilon_S),\beta^\prime_1(\epsilon_1),\ldots, \beta^\prime_1(\epsilon_S)$  \\ 
        \hline
    \end{tabular}%
  \label{tab:dataExplanation}%
\end{table}%

\subsection{Setting for Feature Time Series }
\label{sec:time}

\begin{figure}[h]
\centering
  \includegraphics[width=0.8\linewidth]{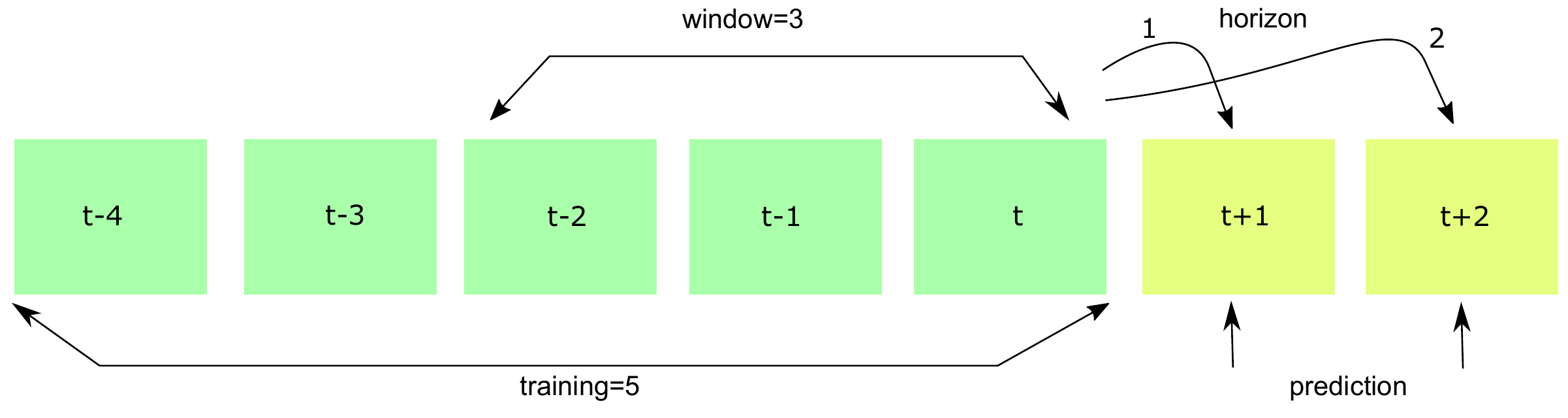}
  \caption{The sliding window based regressor model. The example model trains with data from the last $m=5$ days, and uses the data from $t,~t-1~\text{and}~t-2$ (window=3) to make a prediction for either day $t+1$ (horizon=1) or day $t+2$ (horizon=2).}
  \label{fig:slidingwindow}
\end{figure}

Given the features, we employ a time based approach to predict the Bitcoin price, as shown in Figure~\ref{fig:slidingwindow}. Our goal is to catch trends in the price data, based on the observation that price movements in the preceding days are a good indicator of future prices.   

ChainNet employs three time related concepts: training length, window (lag) and horizon.  Training length is the number of past time periods whose data we use to train our model. Window is the number of past time periods whose data we use to predict Bitcoin price. Horizon is the number of days whose price we predict ahead. 

In the most basic case of prediction horizon $h=1$ and prediction window $w=1$, the model learns to predict the price of day $\hat{y}_{t+1}$ by using the data $x_t$ of day $t$. Similarly, for any window $w$, the model uses data from $\{x_{t-w},\ldots,x_t\}$ to predict the price $\hat{y}_{t+h}$. 

Details of the sliding prediction approach is given in Algorithm~\ref{alg:slidingPrediction}. Input is time indexed data points and output is the model parameters trained on the given input. For given window $w$ and horizon $h$ values, time series data is processed to utilize the history of the current day, $t$ (Line 2-5 in Alg.~\ref{alg:slidingPrediction}). Each $x_{t}$ is replaced by the successive values of time series between $t-w-h$ and $t-h$ (Line 3 in Alg.~\ref{alg:slidingPrediction}). Newly generated $\hat{x_{t}}$ is and its corresponding price, $y_{t}$, is appended to the train list (Line 4-5 in Alg.~\ref{alg:slidingPrediction}).
After all days are iterated on, dimension reduction is applied to the generated $\hat{x}_{train}$ to obtain compensated data (Line 6 in Alg.~\ref{alg:slidingPrediction}). At the end, model is optimized with the previously obtained train data and the algorithm returns the obtained model parameters for out-of-sample predictions (Line 7-8 in Alg.~\ref{alg:slidingPrediction}).

\begin{algorithm}
\footnotesize
    \caption{SPred: Sliding prediction}
    \label{alg:slidingPrediction}
    \begin{algorithmic}[1]
        \Require Data:$\{\left (x_t,y_t\right)$: $ t\in T\}$
        where $x_{t}\in \mathbb{R}^d$; $y_{t}$: the daily bitcoin price in dollars; l: training length; w: sliding window length; h: prediction horizon; $d_2$: pca dimension 
        \Ensure $\theta$: Model Parameters.
            \State $x_{train}, \hat{x}_{train}, y_{train} \leftarrow \{\}$
            \For {each $t\in[h+w:l]$}		
        	   \State $\hat{x}_{t} \gets \left[x_{t-w-h+1},\ldots, x_{t-h}; y_{t-w-h+1},\ldots, y_{t-h}\right]$ // row-wise
        	   \State $\hat{x}_{train} \leftarrow \hat{x}_{train} \cup \hat{x}_{t}$ 
        	   \State $y_{train} \leftarrow y_{train} \cup \hat{y}_{t}$
			\EndFor
        \State $x_{train} \gets PCA(d_2, \hat{x}_{train})$ \label{line:pca}
		\State $\theta =$ model.fit($x_{train}$, $y_{train}$)
	\State \Return{$\theta$}
    \end{algorithmic}  
\end{algorithm}

 We consider the following two parameters in all predictive models: window $w \in \left\{3,5,7\right\}$, horizon $h \in \left\{1,2,5,7,10,15,20,25,30\right\}$, training length $l \in \left\{25,50,100,200\right\}$. As the interaction of horizon, window and training length  parameters may exhibit nonlinear effects on the prediction, we conduct a grid search by varying all parameters, and report the predicted price values for the best model.  

An important point in our sliding prediction approach is that, we train a model per each prediction. As a result, we train a model 365 times to predict Bitcoin prices in 2017. We chose this setting because gain results improved over a batch prediction model. As we model data with low dimensional features, the cost of this approach was negligible.

\subsection{Statistical and Machine Learning Models} 
We evaluate ChainNet performance  by using one statistical and four machine learning models:

\textbf{ARIMAX} refers to the Auto-Regressive Integrated Moving Average model (with exogeneous variable) that is a conventional benchmark model in time series analysis and forecasting that accounts for data non-stationarity~\cite{box2015time}.

\textbf{XGBT} is the eXtreme Gradient Boosting which applies gradient boosting algorithms to decision trees~\cite{chen2016xgboost}.
  
\textbf{RF} stands for Random Forest which is a supervised ensemble of multiple simple decision trees to estimate the dependent variables of the data~\cite{tin1995RF}.

\textbf{GP} presents Gaussian Process based Regression technique which is designed to estimate the regressor parameters with the maximum likelihood principle~\cite{williams1996gaussian}.

\textbf{ENET} refers to the elastic net model which is designed as a regularized linear regression model with the L1 and L2 penalties of the \textit{lasso} and \textit{ridge} methods~\cite{zou2005regularization}.
 
\paragraph{Deep Learning Models.} Given the recent popularity of Deep Learning (DL), we also considered Recurrent Neural Networks and Long Term Short Memory models in ChainNet. However, our experiments did not yield satisfactory results. We hypothesize that DL requires more training data to achieve convergence than we can possibly supply at this point.

\paragraph{Parameter Setting for Models.} For the hyper-parameter tuning of ARIMAX, the orders for auto-regression and moving average terms are chosen from $\left\{0, 1, 2\right\}$. For the tree based approaches such as XGBT, RF, generated number of trees are chosen from  \{10, 50, 100, 200, 300, 400, 500, 1000\}. For the learning rate of XGBT, we tried values from $\left\{0.01, 0.1, 1.0\right\}$. ENET regularization parameters for L1 and L2 and penalty constants are selected from \{0.0001, 0.001, 0.01, 0.1, 1.0, 10.0\} and \{0.001, 0.005, 0.01, 0.05, 0.1, 0.5, 1.0\}. In hyper-parameter tuning of GP, regression types, correlation types, and regularization parameters are chosen from $\{\text{constant, linear, quadratic}\}$, \{absolute exponential, squared exponential, generalized exponential, cubic, linear\}, \{0.001, 0.01, 0.1, 1.0, 10.0\} respectively.

\paragraph{High Dimensionality.} Since we use a windowed (lagged) history of the data, dimensionality of the training data increases rapidly. 

For example, consider the Betti model with $S=50$ filtrations. In addition to Price and TotalTx, each day has $50$ $\beta_0$ and $50$ $\beta_1$ Betti values. For $w=3$, the model uses $(3\cdot(100+2)=306)$ features, whereas there can be at most $(2018-2009)*365$ training instances if we use the entire Bitcoin history. Decreasing the number of scales (e.g., $S=5$) can reduce dimensionality, but this approach reduces the power of Betti models as well, due to decreased threshold granularity. 

We ameliorate the effects of high dimensionality by applying Principal Component Analysis (PCA~\cite{jolliffe2011principal}) to the lagged feature sets of FL, Betti and Betti derivative; in Algorithm~\ref{alg:slidingPrediction} Line \ref{line:pca} PCA maps the high dimensional data into low dimensional data with the dimension of $d_2 \in \left\{5,10,15,20\right\}$.  
 
\subsection{Baseline Performance}

The simplest baseline for ChainNet can be constructed by training models on Price and TotalTx in a sliding window prediction scheme. 
We did not use other baseline features such as mean degree (see discussion in Section~\ref{sec:dataset}) since adding those features reduces the performance of the baseline models.
We train baseline models without reducing the dimensionality ($d_2$=d in Alg.~\ref{alg:slidingPrediction}), because input features are very few; for $w=3$, the models use 6 features in training. We assess model performance with root mean squared error (RMSE) as follows:
 
	$RMSE = \sqrt {{{1}/{\left| T\right|}} {\sum\limits_{t \in T} (y_{t}-\hat{y}_{t})^2}}$,
 where \textit{$\left| T\right|$} is the number of days, $\hat{y}_t$ is the predicted price and $y_t$ is the true observed price on the $t^{th}$ day.

 \begin{figure*}[ht]
 \centering
\begin{subfigure}{0.3\textwidth}
 \includegraphics[width=1.0\linewidth]{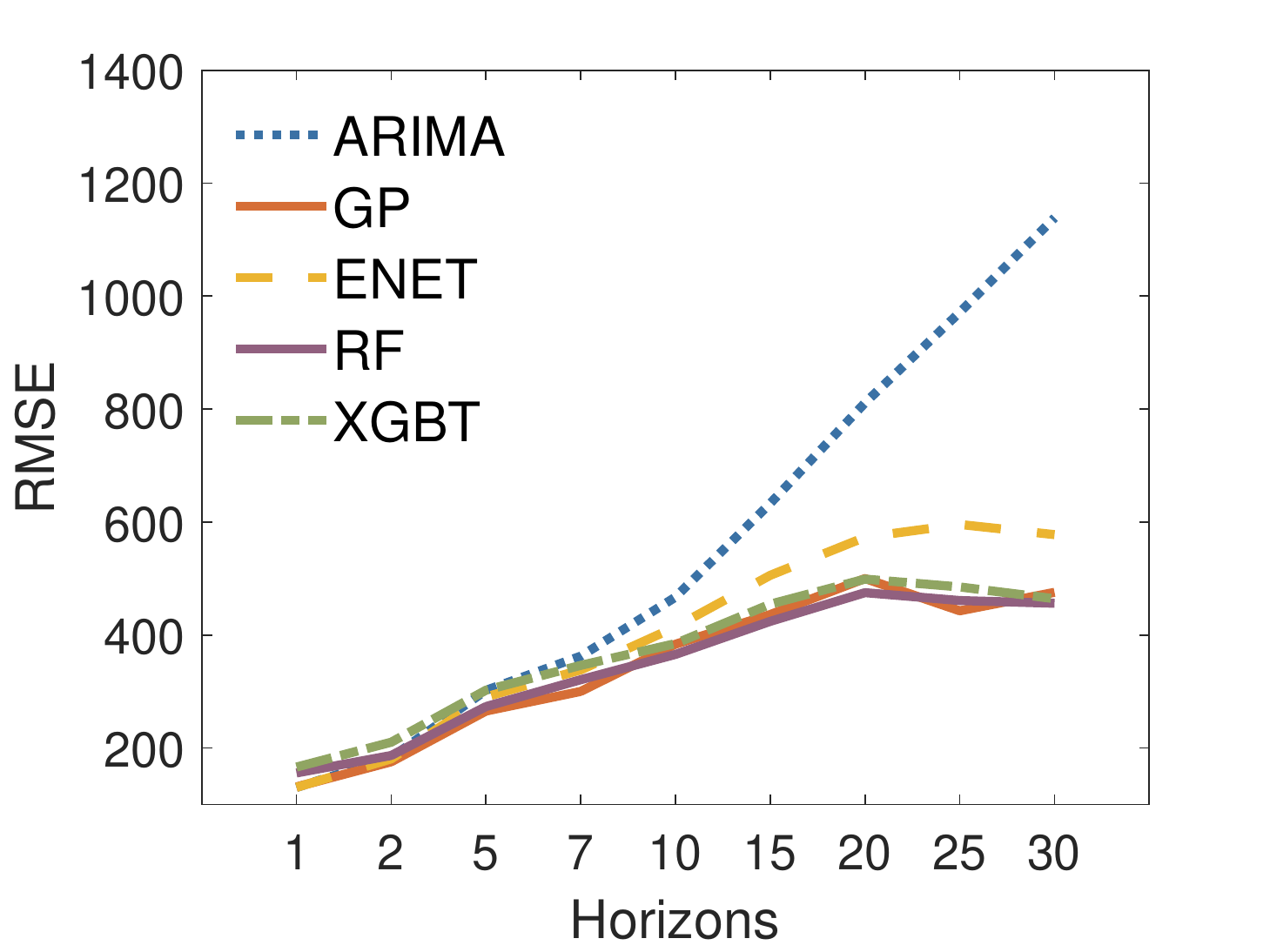}
  \caption{Window=3}
  \label{fig:baseline3}
\end{subfigure}%
~
\begin{subfigure}{0.3\textwidth}
   \includegraphics[width=1.0\linewidth]{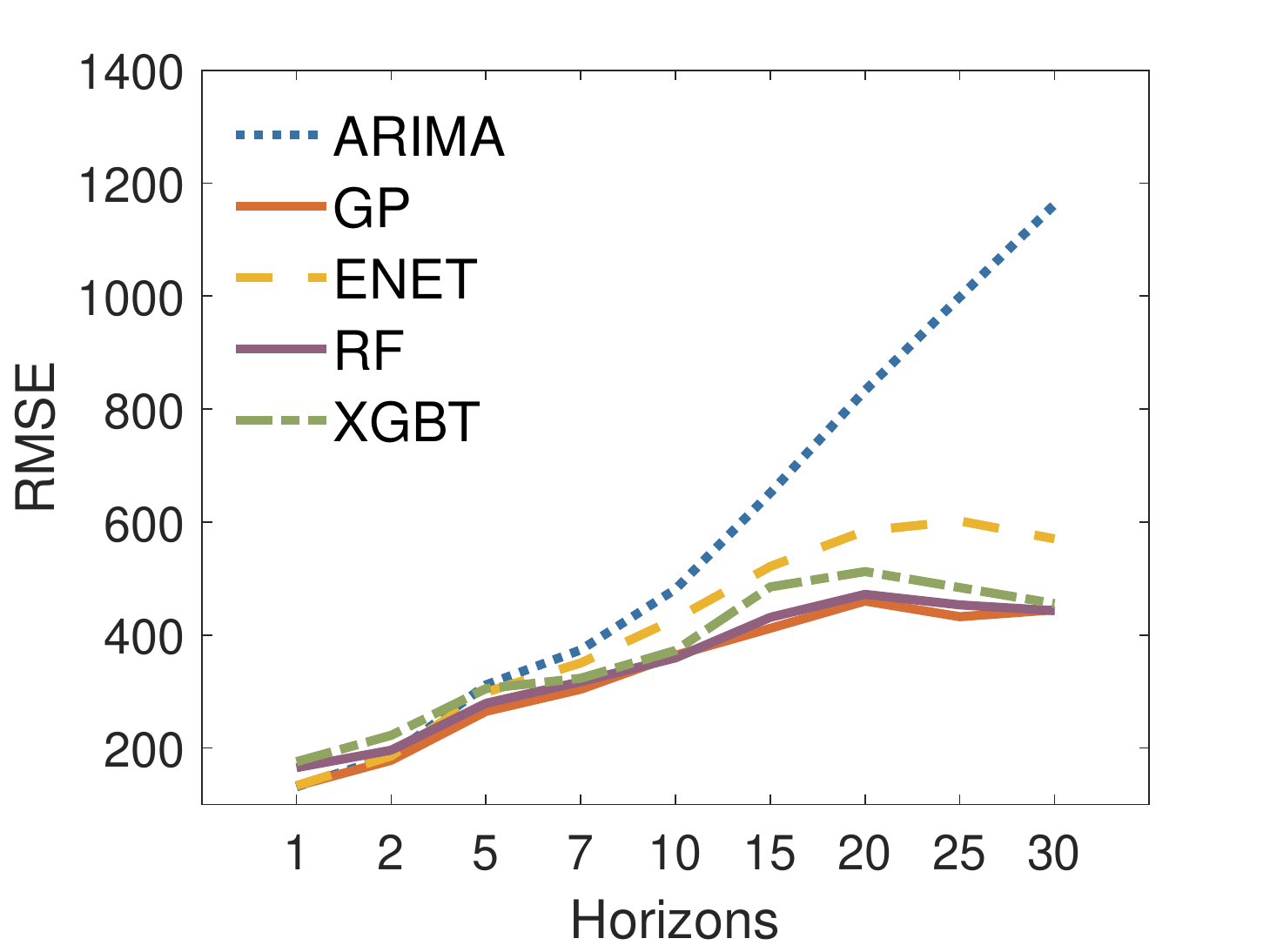}
  \caption{Window=5}
  \label{fig:baseline5}
\end{subfigure}%
~
\begin{subfigure}{0.3\textwidth}
   \includegraphics[width=1.0\linewidth]{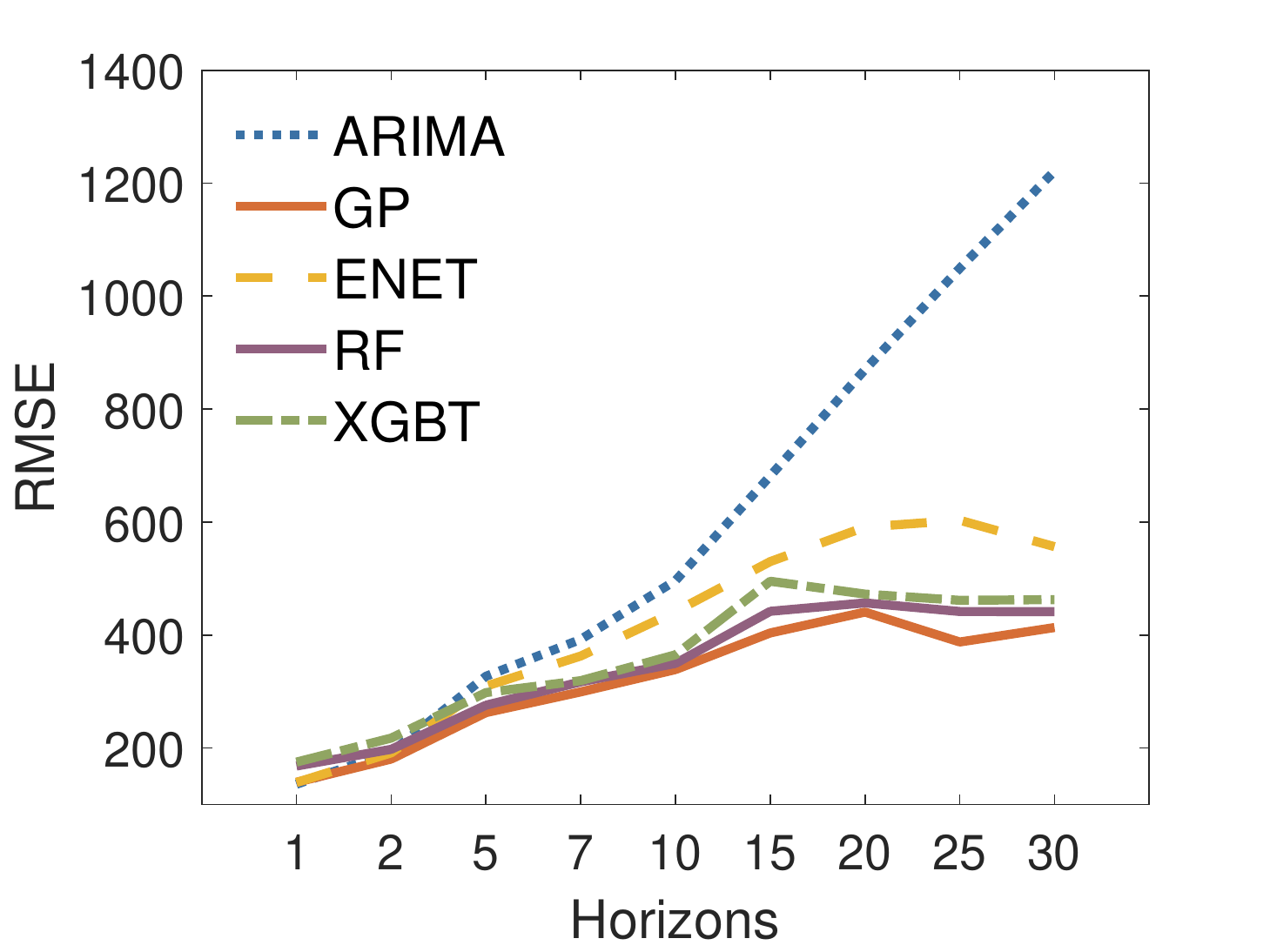}
  \caption{Window=7}
  \label{fig:baseline7}
\end{subfigure}
\caption{RMSE of sliding window based predictions of 2017 Bitcoin prices in different window and horizon values.}
\label{fig:resbaseline}
\end{figure*}

In our rolling predictive framework, we achieve the best results with a training length of 100 days, that is, each considered model is adaptively re-estimated for each $y_{t}$ using data from the previous 100 days. We only report the best results from each model with the hyper-parameter optimization.

Figure~\ref{fig:resbaseline} shows the performance of the five models in prediction. ARIMAX has the worst performance for $h>7$, whereas Gaussian Process (GP) has the best RMSE values overall. We note that as the window value increases, performance does not improve. This implies that considering past information on price and total number of transactions does not deliver improvement in forecasting accuracy. In fact, from window 3 to 7, the RMSE values of the best model, GP, is approximately similar while $h<10$. For $h>10$, the RMSE values decrease 13\% from window 3 to 7.

\paragraph{Other Baseline Research Studies}
We use the results of~\cite{akcora2018chainlet} as a baseline comparison for ChainNet. 
The maximum gain achieved by~\cite{akcora2018chainlet} over the models without chainlets is 12.5\% at forecasting horizon of 30 days; in turn,
the highest gain of ChainNet for the same horizon of $h=30$ is approximately 20\%, that is, 7.5\% improvement of ChainNet over~\cite{akcora2018chainlet}.  Furthermore,
the highest gain of ChainNet among all forecasting horizons is approximately 40\% and is achieved at $h=15$, that is, more than three times improvement over the highest gain of~\cite{akcora2018chainlet}. 
 
Finally, the closest scholarly work to ChainNet is detailed in a report by Greaves et al.~\cite{greaves2015using}, where the authors extract both graph centric features (e.g., mean degree) and transaction features (e.g., mean amount) from the Bitcoin address graph, and use support vector machines to predict the Bitcoin price. As the authors also note at the end of their study, these features do not bring more information over a model that uses price data only. Indeed our experiments showed high error rates for predictions with the authors' experimental setting. More powerful models have been used in \cite{kondor2014inferring,shah2014bayesian} with better results. We adopt similar machine learning models in this work, but in addition to the traditional features (see Table~\ref{tab:dataExplanation}) ChainNet utilizes novel feature sets in FL, Betti and Betti derivative models.

\subsection{ChainNet Model Performance}

In this section, we provide performance of the predictive models built with FL, Betti and Betti derivative features.  \textit{Our hypothesis is that adding these features will increase model performance}, i.e., RMSE in predictions will decrease  over their associated baseline values. 

\paragraph{Performance Gain}
In our analysis, we report the percentage predictive gain, or decrease in $RMSE$ for a specific machine learning model $m$ w.r.t. its baseline model $m_0$ as $\Delta_{m}(w,h)= 100\times \bigl(1 - {RMSE_m(w,h)}/{RMSE_{m_0}(w,h)}\bigr)$,
where $RMSE_{m_0}(w,h)$ and $RMSE_m(w,h)$ are delivered by a baseline model $m_0$ and a competing model $m$, respectively.

 \begin{figure*}[ht]
 \centering
\begin{subfigure}{0.30\textwidth}
 \includegraphics[width=1.0\linewidth]{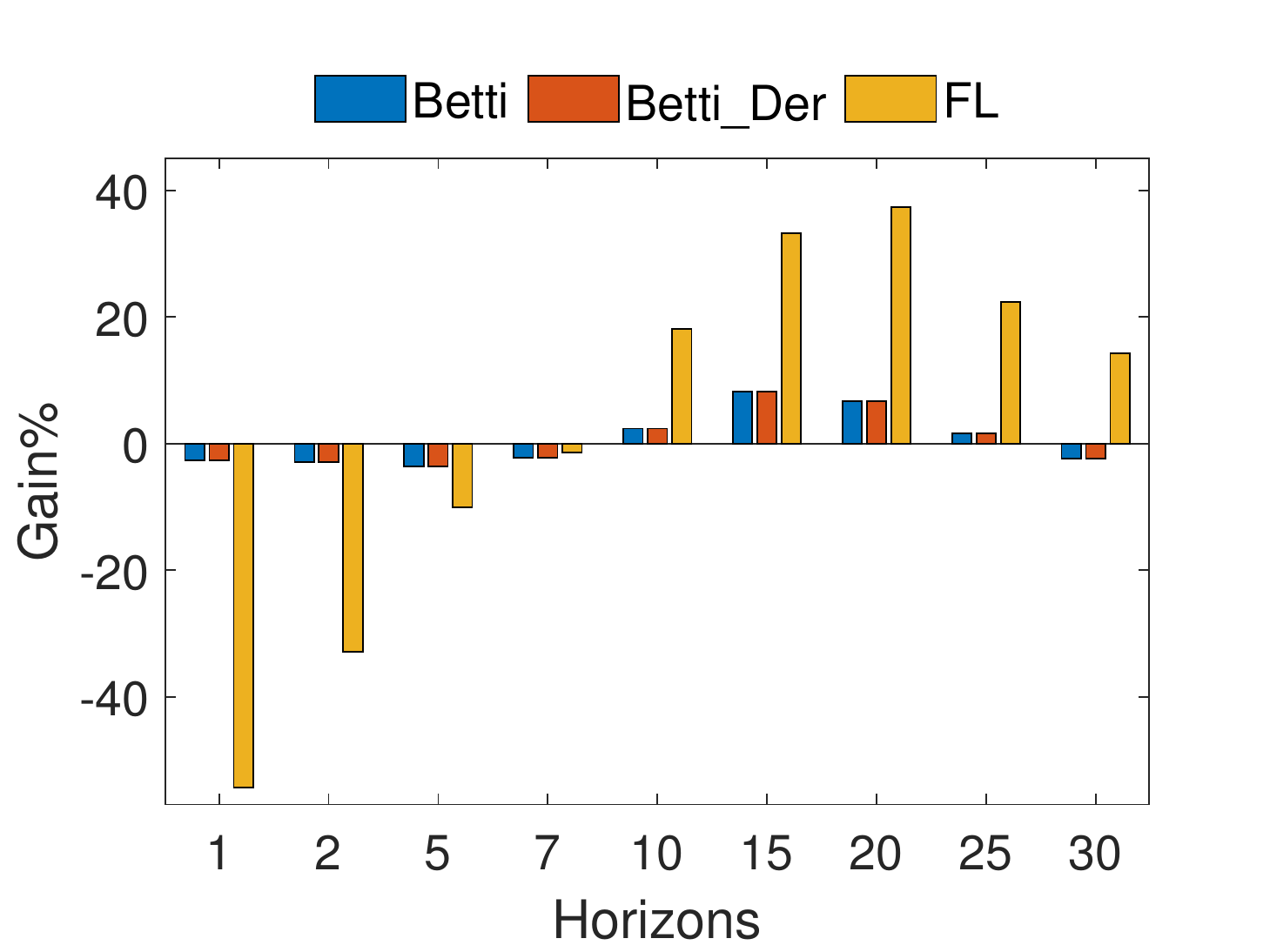}
  \caption{Window=3}
  \label{fig:enet_gain_w3}
\end{subfigure}%
~
\begin{subfigure}{0.30\textwidth}
   \includegraphics[width=1.0\linewidth]{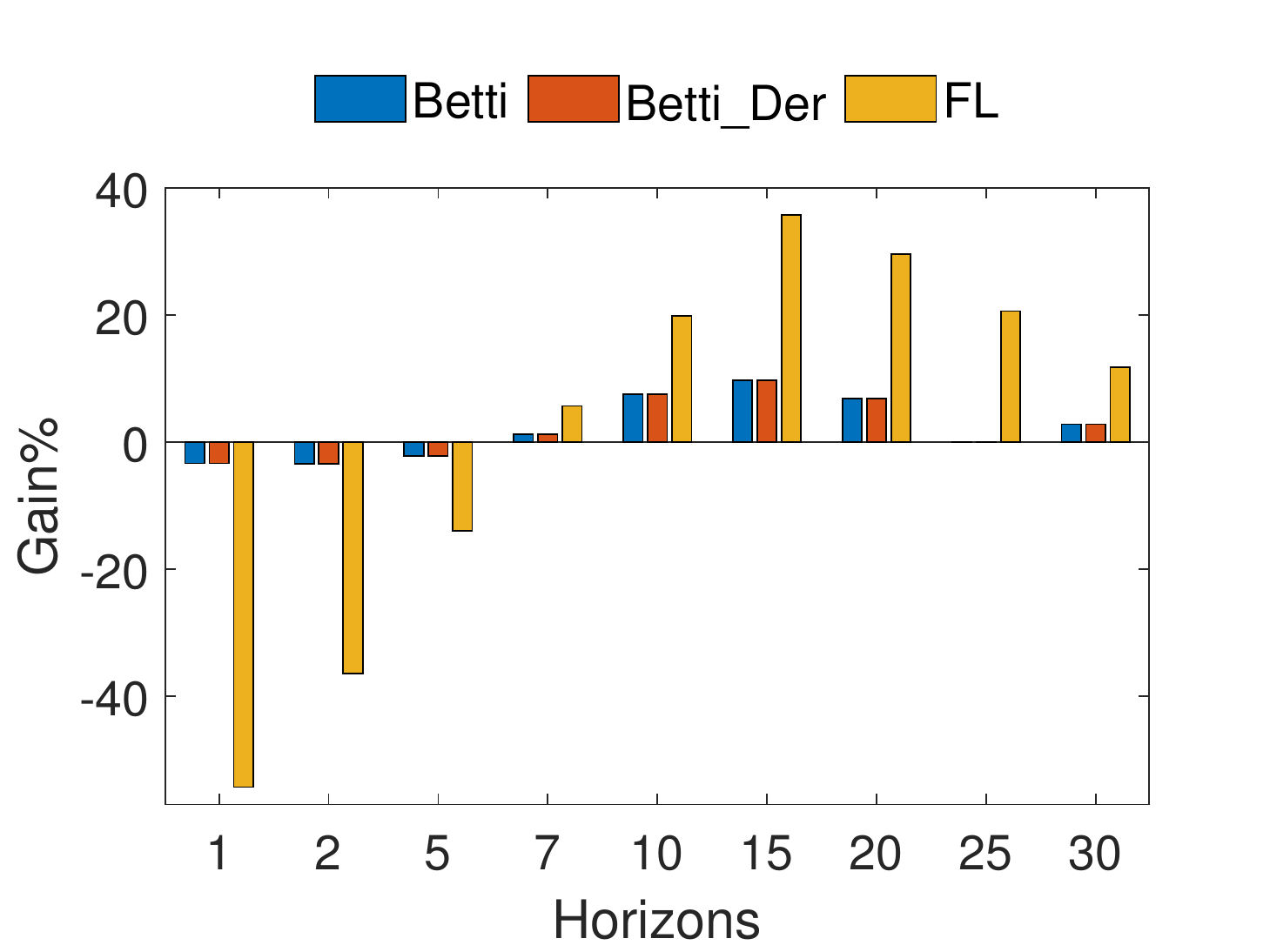}
  \caption{Window=5}
  \label{fig:enet_gain_w5}
\end{subfigure}%
~
\begin{subfigure}{0.30\textwidth}
   \includegraphics[width=1.0\linewidth]{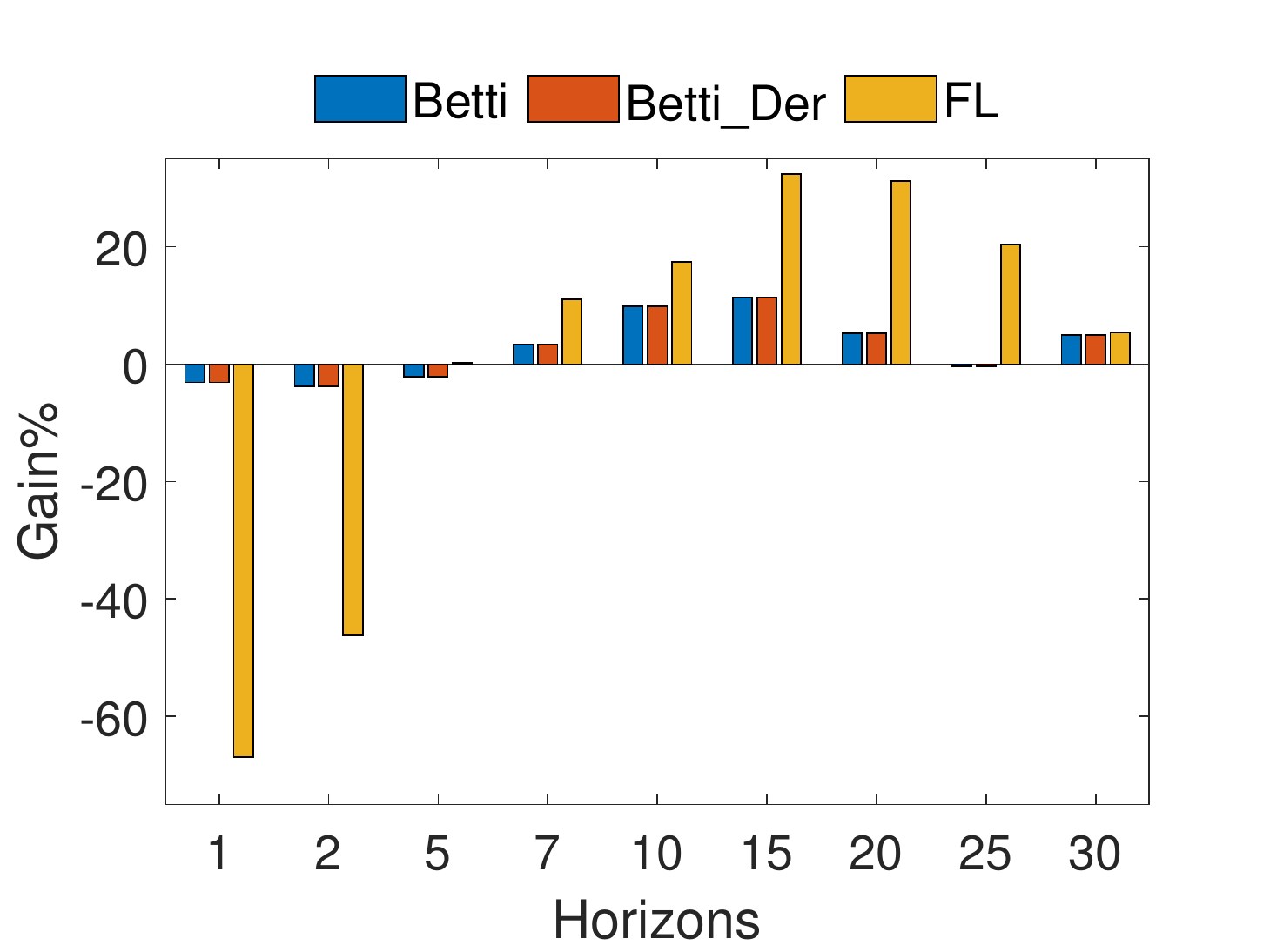}
  \caption{Window=7}
  \label{fig:enet_gain_w7}
\end{subfigure}
\caption{Elastic Net model performance.}
\label{fig:enet}
\end{figure*}

\begin{figure*}[ht]
 \centering
\begin{subfigure}{0.3\textwidth}
 \includegraphics[width=1.0\linewidth]{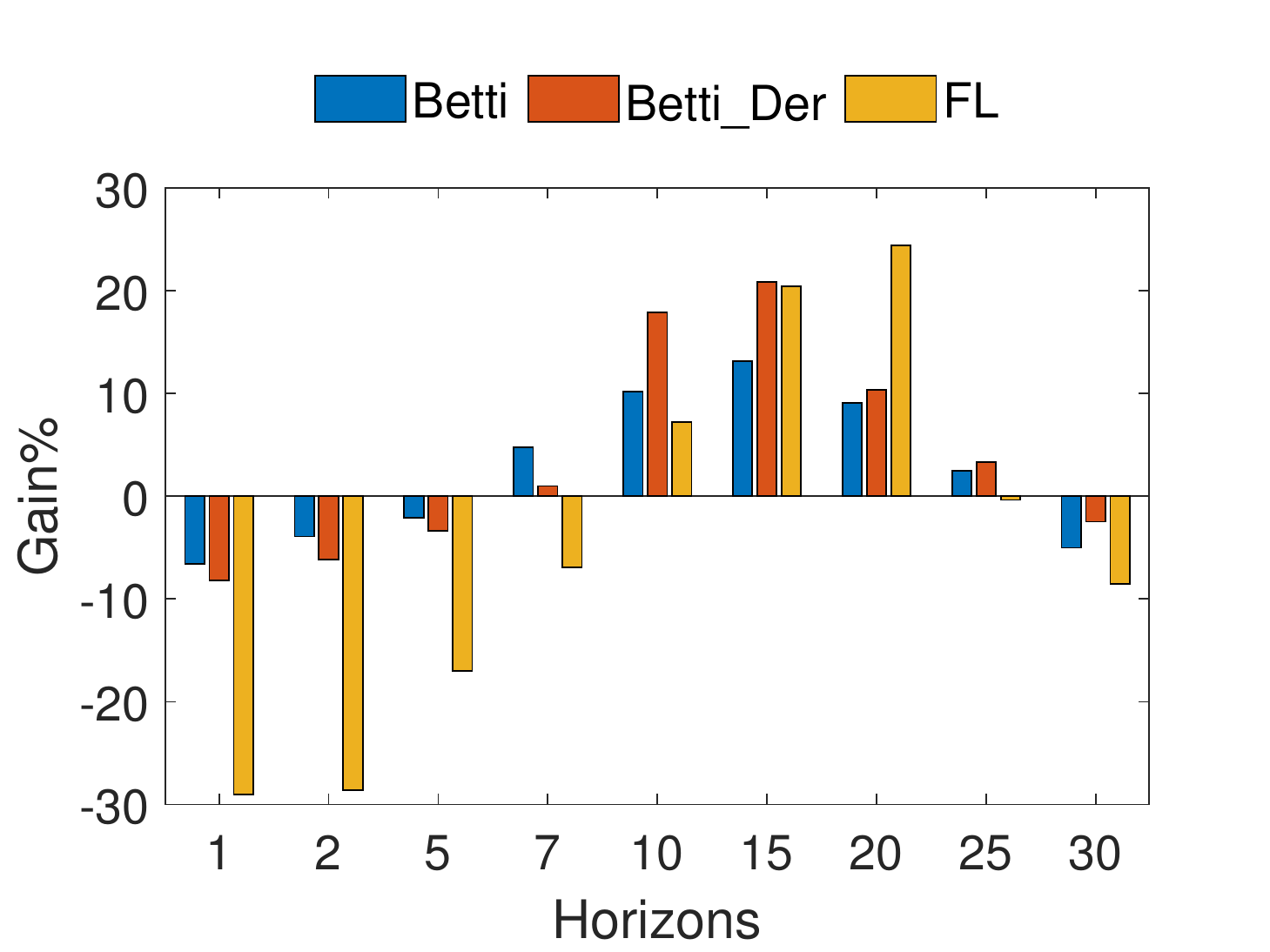}
  \caption{Window=3}
  \label{fig:rf_w3_gain}
\end{subfigure}%
~
\begin{subfigure}{0.30\textwidth}
   \includegraphics[width=1.0\linewidth]{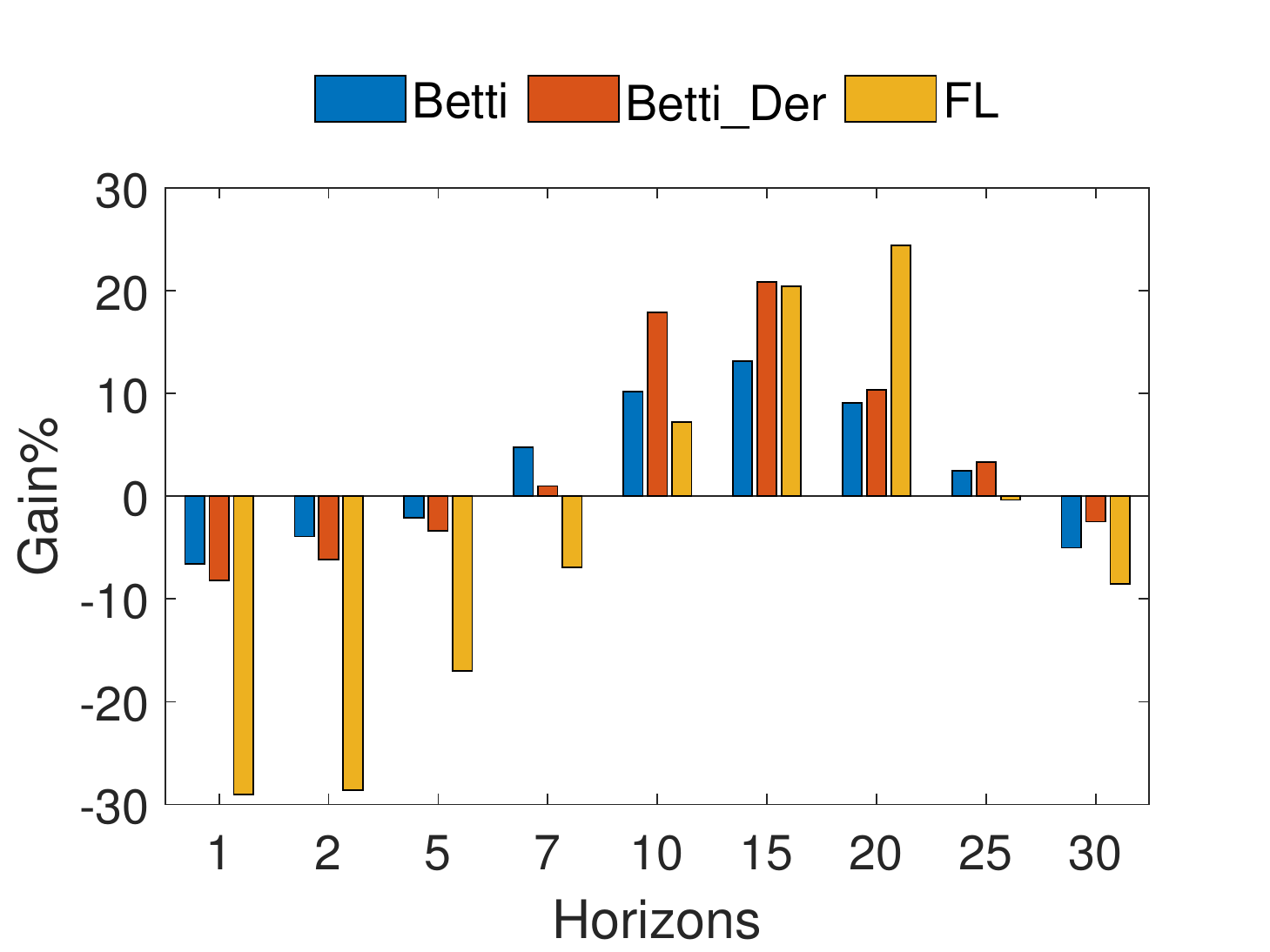}
  \caption{Window=5}
  \label{fig:rf_w5_gain}
\end{subfigure}%
~
\begin{subfigure}{0.3\textwidth}
   \includegraphics[width=1.0\linewidth]{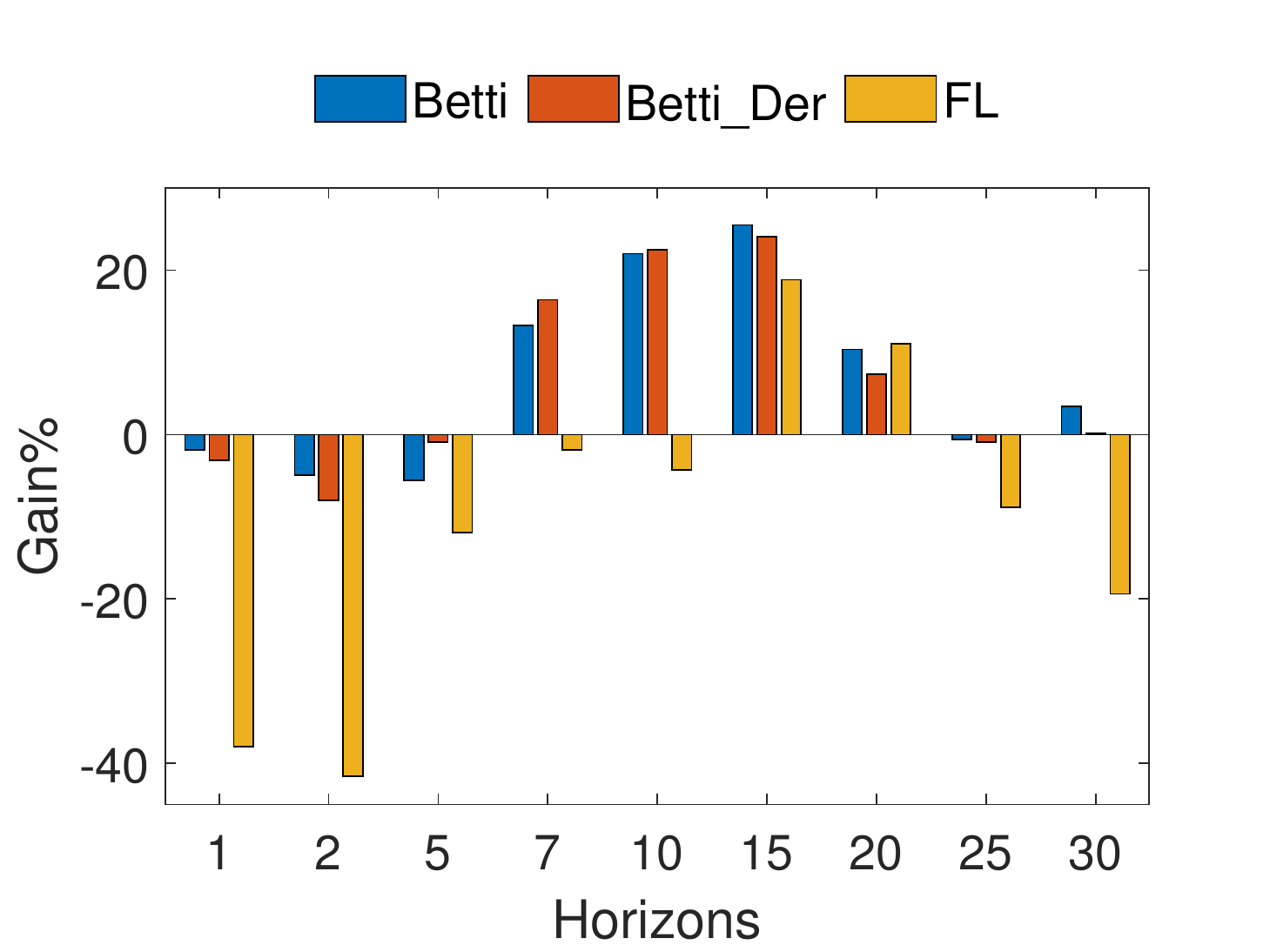}
  \caption{Window=7}
  \label{fig:rf_w7_gain}
\end{subfigure}
\caption{Random Forest Performance.}
\label{fig:rf}
\end{figure*}

Enet performance results are shown in Figure~\ref{fig:enet}, which indicate that up to seven days, models do not improve when trained with ChainNet features. A similar trend is visible in the Random Forest(RF) results, as given in Figure~\ref{fig:rf}. However, in RF results, for increasing horizons gain values dip below 0\%,  whereas Enet gains stay above 0\%. In both models $h=1,\ldots,5$ predictions have negative gains. These results indicate that for immediate future, these machine learning models perform better \textit{when trained on price and transaction counts (TotalTx) only}. 

Intuitively, if Bitcoin price increases/decreases consistently in the last $w$ days, we expect the trend to continue in the following days. RF and ENET models capture this trend better without the ChainNet features in short horizons.  

\begin{figure*}[ht]
 \centering
\begin{subfigure}{0.3\textwidth}
 \includegraphics[width=1.0\linewidth]{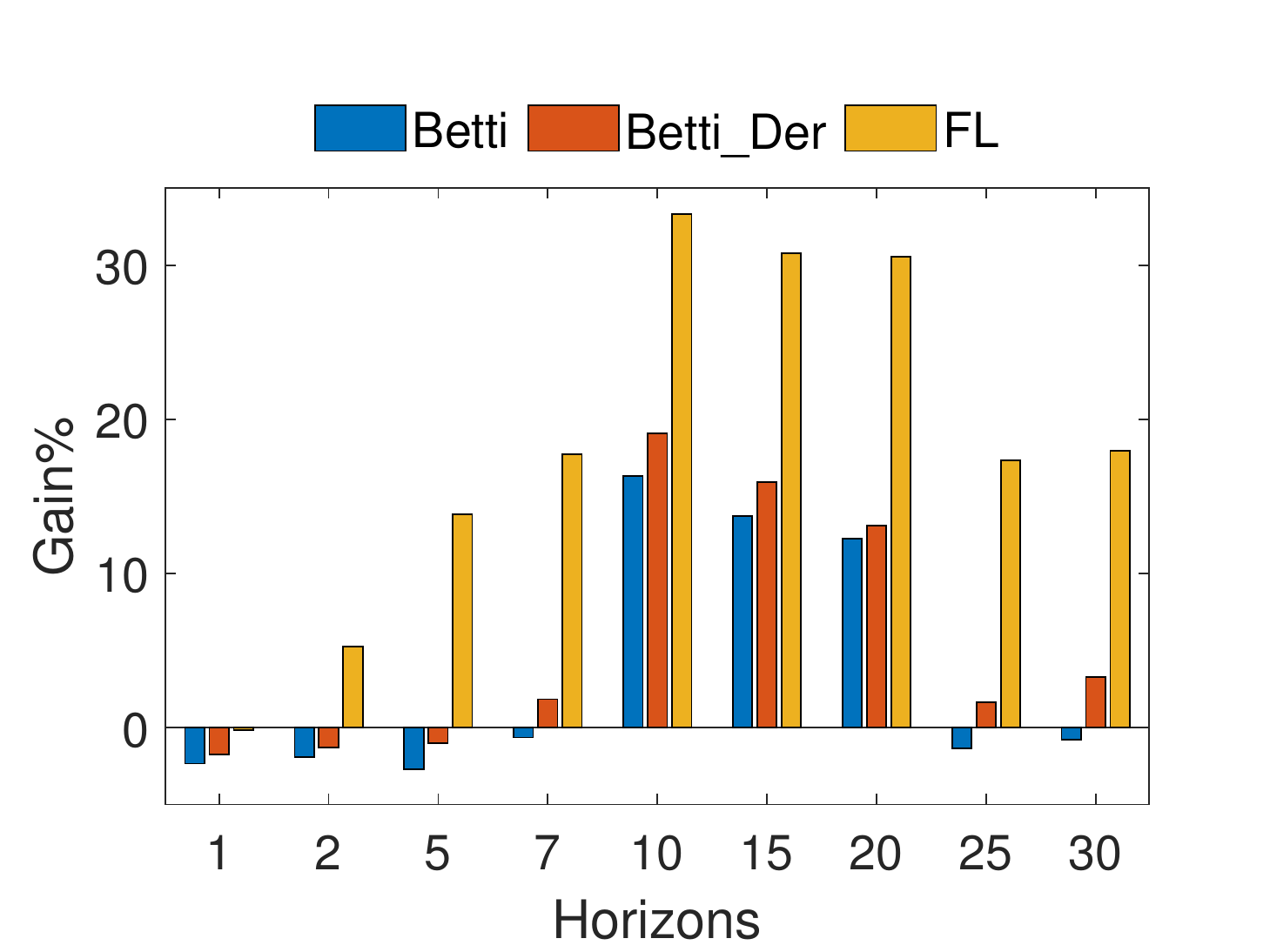}
  \caption{Window=3}
  \label{fig:gp_w3_gain}
\end{subfigure}%
~
\begin{subfigure}{0.3\textwidth}
   \includegraphics[width=1.0\linewidth]{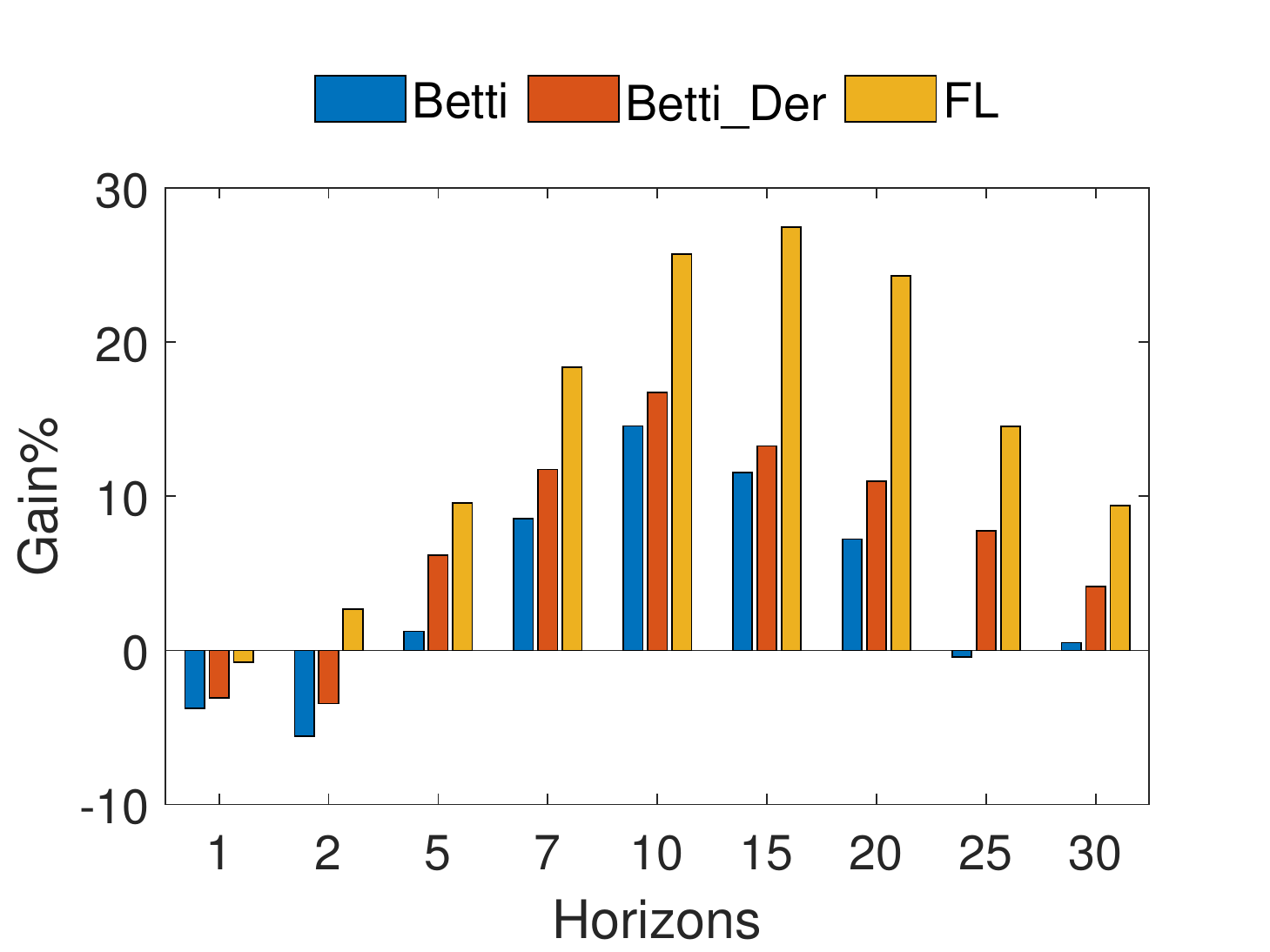}
  \caption{Window=5}
  \label{fig:gp_w5_gain}
\end{subfigure}%
~
\begin{subfigure}{0.3\textwidth}
   \includegraphics[width=1.0\linewidth]{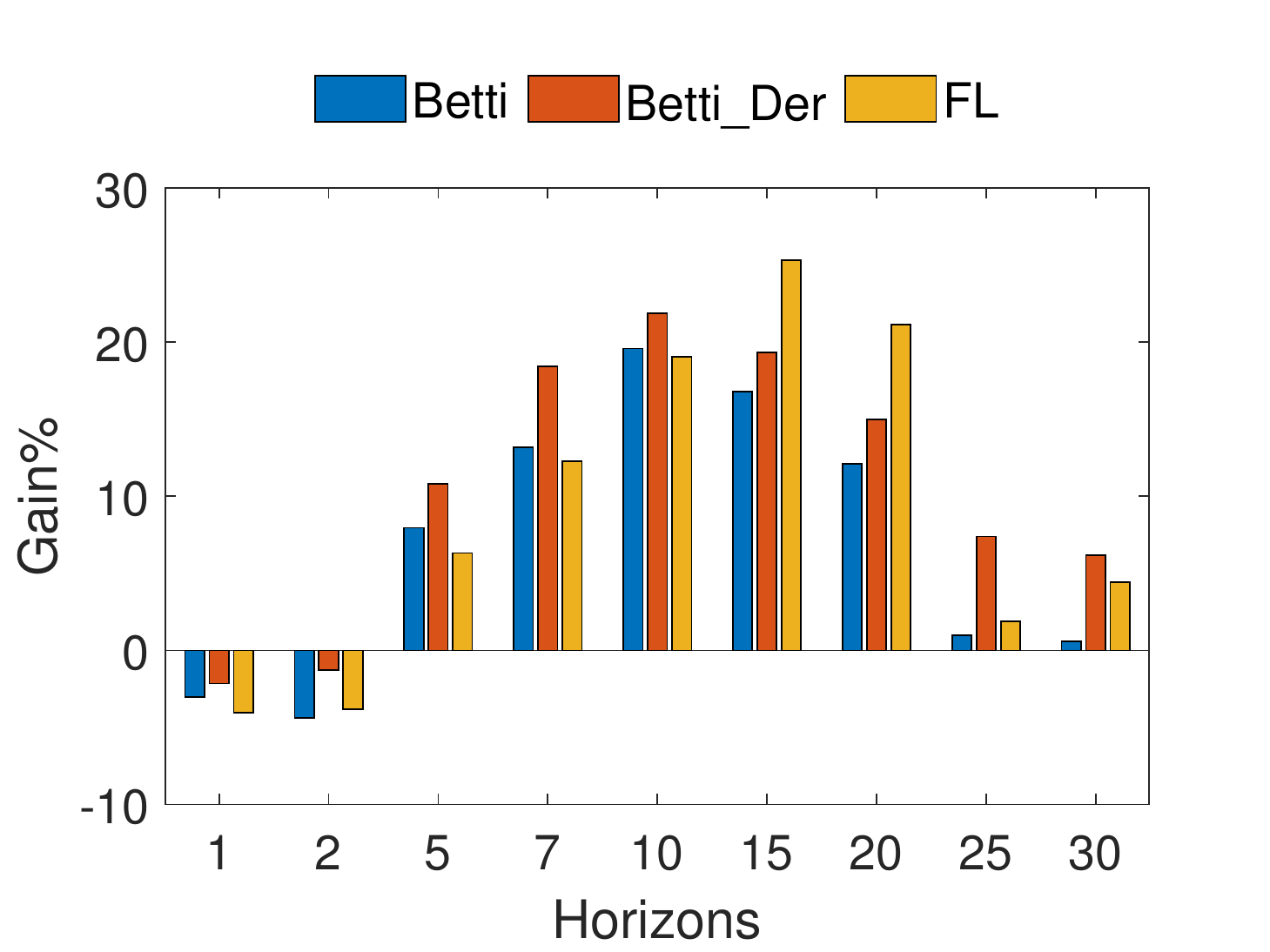}
  \caption{Window=7}
  \label{fig:gp_w7_gain}
\end{subfigure}
\caption{Gaussian Process (GP) based regression performance. }
\label{fig:gp}
\end{figure*}

\begin{figure*}[ht]
 \centering
\begin{subfigure}{0.3\textwidth}
 \includegraphics[width=1.0\linewidth]{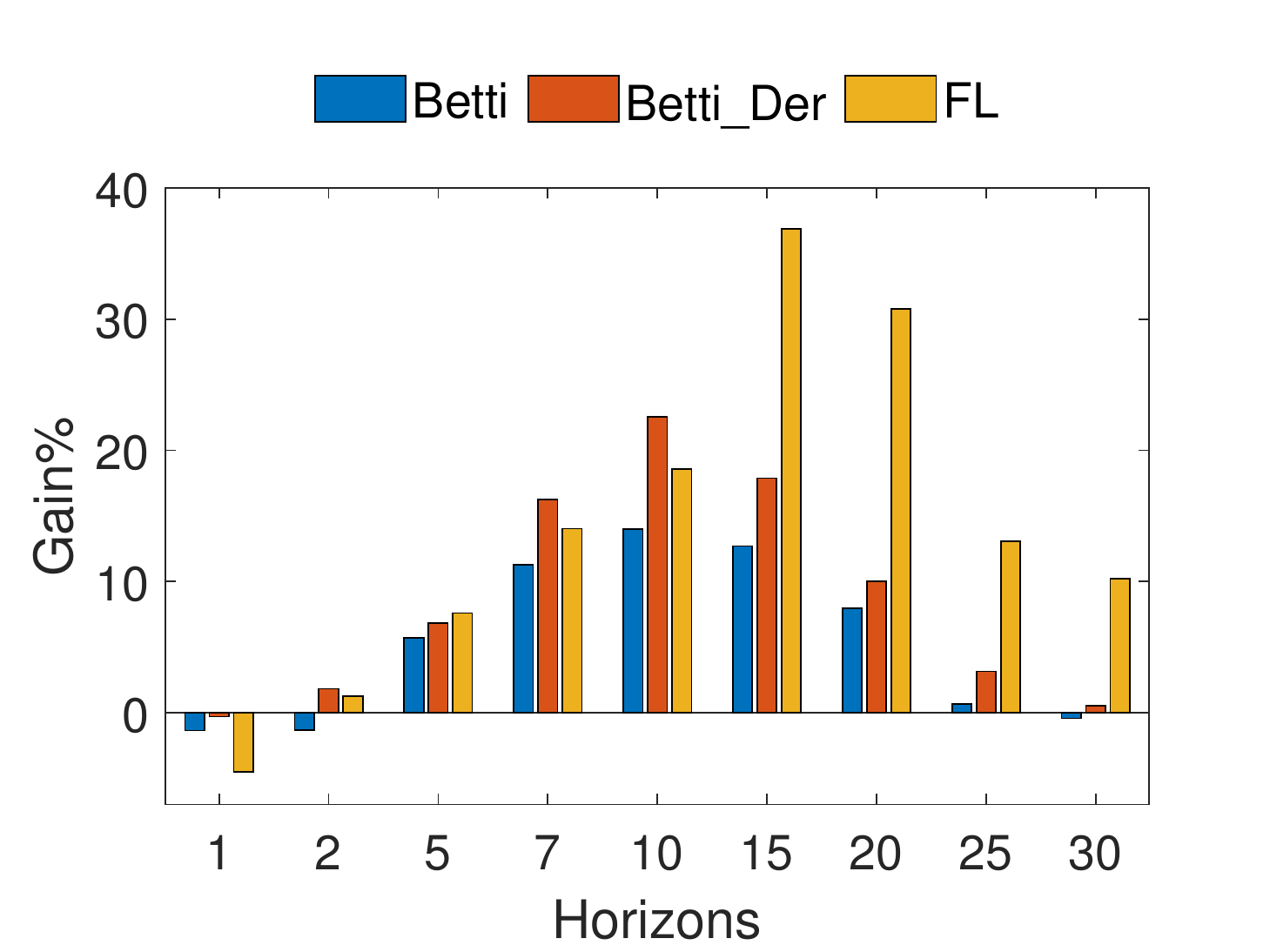}
  \caption{Window=3}
  \label{fig:xgbt_w3_gain}
\end{subfigure}%
~
\begin{subfigure}{0.3\textwidth}
   \includegraphics[width=1.0\linewidth]{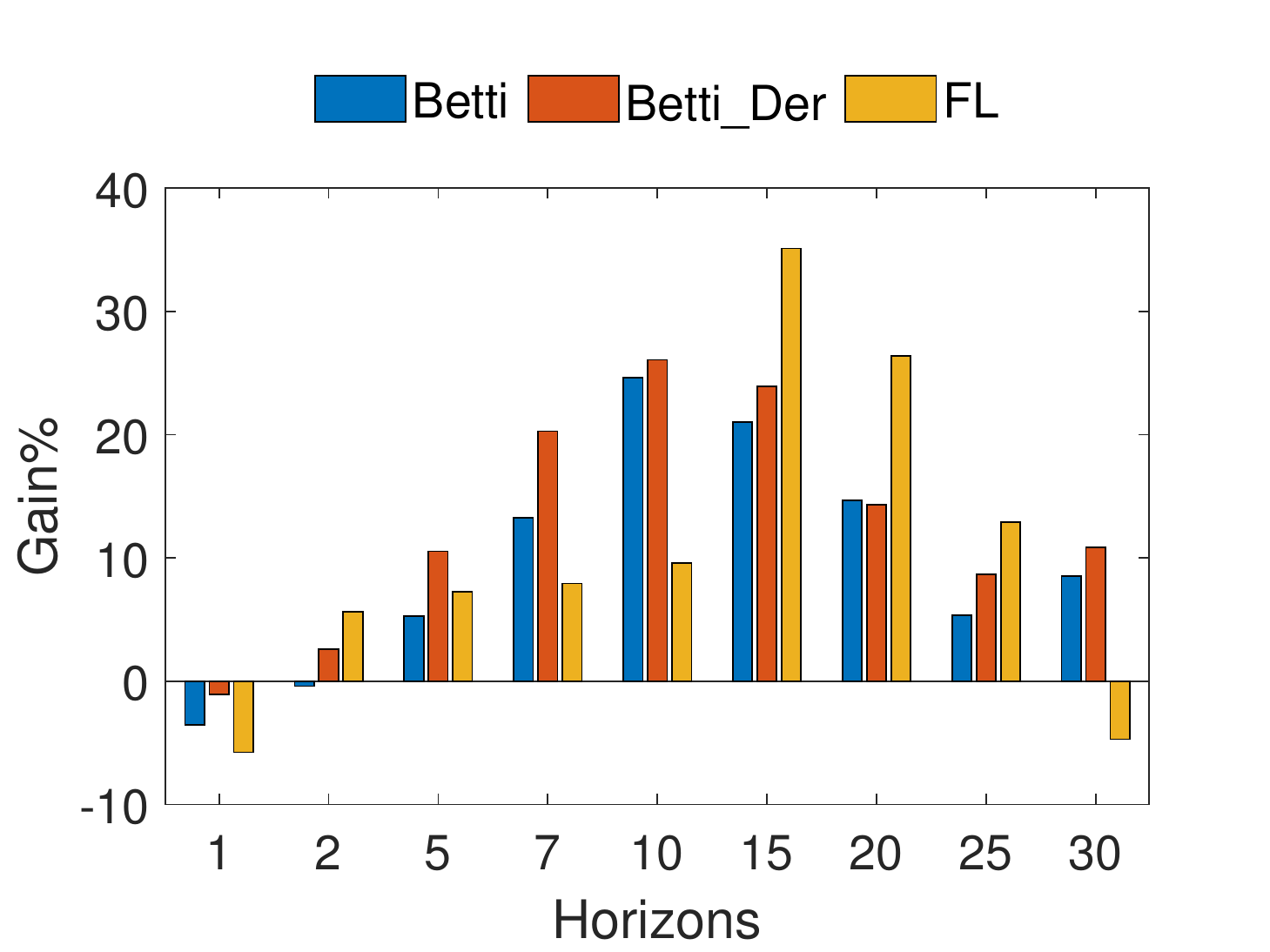}
  \caption{Window=5}
  \label{fig:xgbt_w5_gain}
\end{subfigure}%
~
\begin{subfigure}{0.3\textwidth}
   \includegraphics[width=1.0\linewidth]{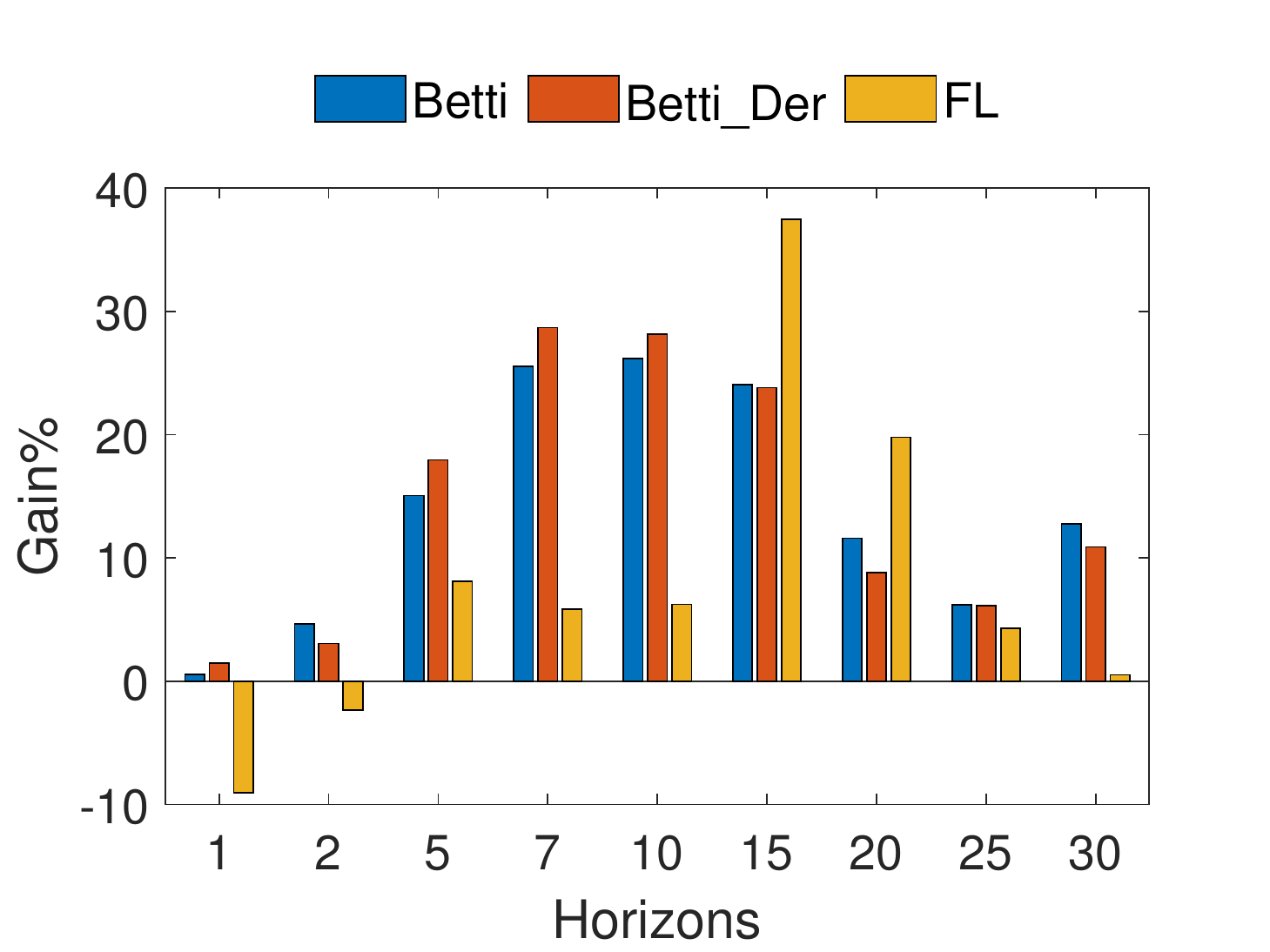}
  \caption{\textcolor{blue}{Best gain results: Window=7}}
  \label{fig:xgbt_w7_gain}
\end{subfigure}
\caption{ Extreme  Gradient  Boosting (XGBT) performance.}
\label{fig:xgbt}
\end{figure*}

Figures~\ref{fig:gp} and \ref{fig:xgbt} show that XGBT and GP predictions improve for increasing horizons, but decrease for $h>15$. Specifically $h=1$ predictions reach a positive gain only in XGBT $w=7$. XGBT also offers the best gains for $h=2$, but its performance deteriorates for $h>15$. 

In constructing the XGBT model, the boosting approach focuses on examples that increase the error rate of objective function at each step.  We hypothesize that this specific focus is the reason for XGBT's better performance. 

The highest gain values for $h \leq 7$ are achieved in XGBT Betti models for $w=7$  (38\% in Figure~\ref{fig:xgbt_w7_gain}). Our heuristic approach, FL, has an interesting trend; its usage in models lead to better gains for higher horizons.  On the other hand Betti models achieve better gain values for short horizons. Considering these results, ChainNet can use Betti and Betti derivatives for short $(h<10)$ term prediction, and use FL for $h > 15$. 

An important result is that next day predictions ($h=1$) do not improve significantly (i.e., at most 2\% in Figure~\ref{fig:xgbt_w7_gain}) with ChainNet features. In other words, topological and graph based signals in the blockchain have a negligible causal affect on the next immediate day.

Our results offer evidence for the hypothesis that considering topological features in predictive models bring a significant gain. ChainNet uses Betti models and FL for short and long term predictions, respectively.

\section{Conclusion}
\label{sec:concl}
 
ChainNet is a price prediction platform that utilizes topological characteristics of a blockchain graph. ChainNet builds topological constructs over a graph and computes quantitative summaries in the form of the Betti sequences and Betti derivatives which are then used in model building for the Bitcoin price prediction. Furthermore, ChainNet also offers a heuristic based approach that allows user tailoring of system parameters for a finer grained look. Our results on the full Bitcoin network show that in less than 7 day ahead predictions, Betti models bring a prediction gain of almost 40\% over baseline approaches.

\balance

\bibliographystyle{spbasic}
\bibliography{chartalist.bib}

\begin{thebibliography}{40}
\providecommand{\natexlab}[1]{#1}
\providecommand{\url}[1]{{#1}}
\providecommand{\urlprefix}{URL }
\expandafter\ifx\csname urlstyle\endcsname\relax
  \providecommand{\doi}[1]{DOI~\discretionary{}{}{}#1}\else
  \providecommand{\doi}{DOI~\discretionary{}{}{}\begingroup
  \urlstyle{rm}\Url}\fi
\providecommand{\eprint}[2][]{\url{#2}}

\bibitem[{Akcora et~al.(2018)Akcora, Dey, Gel, and
  Kantarcioglu}]{akcora2018chainlet}
Akcora CG, Dey AK, Gel YR, Kantarcioglu M (2018) Forecasting bitcoin price with
  graph chainlets. In: PAKDD, pp 1--12

\bibitem[{Androulaki et~al.(2013)Androulaki, Karame, Roeschlin, Scherer, and
  Capkun}]{androulaki2013evaluating}
Androulaki E, Karame GO, Roeschlin M, Scherer T, Capkun S (2013) Evaluating
  user privacy in bitcoin. In: IFCA, Springer, pp 34--51

\bibitem[{Antulov-Fantulin et~al.(2018)Antulov-Fantulin, Tolic, Piskorec, Ce,
  and Vodenska}]{antulov2018inferring}
Antulov-Fantulin N, Tolic D, Piskorec M, Ce Z, Vodenska I (2018) Inferring
  short-term volatility indicators from the bitcoin blockchain. In:
  International Workshop on Complex Networks and their Applications, Springer,
  pp 508--520

\bibitem[{Baumann et~al.(2014)Baumann, Fabian, and
  Lischke}]{baumann2014exploring}
Baumann A, Fabian B, Lischke M (2014) Exploring the bitcoin network. In: WEBIST
  (1), pp 369--374

\bibitem[{Box et~al.(2015)Box, Jenkins, Reinsel, and Ljung}]{box2015time}
Box GE, Jenkins GM, Reinsel GC, Ljung GM (2015) Time series analysis:
  forecasting and control. John Wiley \& Sons

\bibitem[{Carlsson(2009)}]{Carlsson}
Carlsson G (2009) Topology and data. Bulletin of American Mathematical Society
  (NS) 46(2):255--308

\bibitem[{{Chazal} and {Michel}(2017)}]{TDAintro}
{Chazal} F, {Michel} B (2017) {An introduction to Topological Data Analysis:
  fundamental and practical aspects for data scientists}. ArXiv e-prints pp
  1--38

\bibitem[{Chen and Guestrin(2016)}]{chen2016xgboost}
Chen T, Guestrin C (2016) Xgboost: A scalable tree boosting system. In: The
  22nd {SIGKDD}, ACM, pp 785--794

\bibitem[{Di~Battista et~al.(2015)Di~Battista, Di~Donato, Patrignani, Pizzonia,
  Roselli, and Tamassia}]{di2015bitconeview}
Di~Battista G, Di~Donato V, Patrignani M, Pizzonia M, Roselli V, Tamassia R
  (2015) Bitconeview: visualization of flows in the bitcoin transaction graph.
  In: IEEE VizSec, pp 1--8

\bibitem[{Dyhrberg(2016)}]{dyhrberg2016bitcoin}
Dyhrberg AH (2016) Bitcoin, gold and the dollar--a garch volatility analysis.
  Finance Research Letters 16:85--92

\bibitem[{Edelsbrunner and Parsa(2014)}]{edelsbrunner2014computational}
Edelsbrunner H, Parsa S (2014) On the computational complexity of betti
  numbers: reductions from matrix rank. In: The 25th ACM-SIAM Symposium on
  Discrete Algorithms, SIAM, pp 152--160

\bibitem[{Filtz et~al.(2017)Filtz, Polleres, Karl, and
  Haslhofer}]{filtzevolution}
Filtz E, Polleres A, Karl R, Haslhofer B (2017) Evolution of the bitcoin
  address graph

\bibitem[{Garg et~al.(2016)Garg, Lu, Popuri, and Beg}]{AmanmeetPD}
Garg A, Lu D, Popuri K, Beg MF (2016) Cortical geometry network and topology
  markers for parkinson’s disease. arXiv preprint:161104393 pp 1--10

\bibitem[{Gionis et~al.(2012)Gionis, Lappas, and Terzi}]{gionis2012estimating}
Gionis A, Lappas T, Terzi E (2012) Estimating entity importance via counting
  set covers. In: The 18th SIGKDD, ACM, pp 687--695

\bibitem[{Greaves and Au(2015)}]{greaves2015using}
Greaves A, Au B (2015) Using the bitcoin transaction graph to predict the price
  of bitcoin. No Data

\bibitem[{Henelius et~al.(2016)Henelius, Ukkonen, and
  Puolam{\"a}ki}]{henelius2016finding}
Henelius A, Ukkonen A, Puolam{\"a}ki K (2016) Finding statistically significant
  attribute interactions. arXiv preprint arXiv:161207597

\bibitem[{Ho(1995)}]{tin1995RF}
Ho TK (1995) Random decision forests. In: Proceedings of 3rd International
  Conference on Document Analysis and Recognition, vol~1, pp 278--282 vol.1,
  \doi{10.1109/ICDAR.1995.598994}

\bibitem[{Hofer et~al.(2017)Hofer, Kwitt, Niethammer, and Uhl}]{Hofer2018}
Hofer C, Kwitt R, Niethammer M, Uhl A (2017) Deep learning with topological
  signatures. NIPS pp 1634--1644

\bibitem[{Hyndman and Fan(1996)}]{hyndman1996sample}
Hyndman RJ, Fan Y (1996) Sample quantiles in statistical packages. The American
  Statistician 50(4):361--365

\bibitem[{Jog and Loh(2015)}]{allertonJogL15}
Jog V, Loh P (2015) Recovering communities in weighted stochastic block models.
  In: 53rd Allerton Conf. on Communication, Control, and Computing, Monticello,
  USA, pp 1308--1315

\bibitem[{Jolliffe(2011)}]{jolliffe2011principal}
Jolliffe I (2011) Principal component analysis. In: International encyclopedia
  of statistical science, Springer, pp 1094--1096

\bibitem[{Kondor et~al.(2014{\natexlab{a}})Kondor, Csabai, Sz{\"u}le, and
  P{\'o}sfai}]{kondor2014inferring}
Kondor D, Csabai I, Sz{\"u}le J, P{\'o}sfai G Mand~Vattay (2014{\natexlab{a}})
  Inferring the interplay between network structure and market effects in
  bitcoin. New J of Phys 16(12):125003

\bibitem[{Kondor et~al.(2014{\natexlab{b}})Kondor, P{\'o}sfai, Csabai, and
  Vattay}]{kondor2014rich}
Kondor D, P{\'o}sfai M, Csabai I, Vattay G (2014{\natexlab{b}}) Do the rich get
  richer? an empirical analysis of the bitcoin transaction network. PloS one
  9(2):e86197

\bibitem[{Kristoufek(2015)}]{kristoufek2015main}
Kristoufek L (2015) What are the main drivers of the bitcoin price? evidence
  from wavelet coherence analysis. PloS {O}ne 10(4)

\bibitem[{Lischke and Fabian(2016)}]{lischke2016analyzing}
Lischke M, Fabian B (2016) Analyzing the bitcoin network: The first four years.
  Future Internet 8(1):7

\bibitem[{Madan and Zhao(2015)}]{madan2015automated}
Madan S Iand~Saluja, Zhao A (2015) Automated bitcoin trading via machine
  learning algorithms

\bibitem[{Mattila et~al.(2016)}]{mattila2016blockchain}
Mattila J, et~al. (2016) The blockchain phenomenon--the disruptive potential of
  distributed consensus architectures. Tech. rep., The Research Institute of
  the Finnish Economy

\bibitem[{Milo et~al.(2002)Milo, Shen-Orr, Itzkovitz, Kashtan, Chklovskii, and
  Alon}]{milo2002network}
Milo R, Shen-Orr S, Itzkovitz S, Kashtan N, Chklovskii D, Alon U (2002) Network
  motifs: {S}imple building blocks of complex networks. Science
  298(5594):824--827

\bibitem[{Nakamoto(2008)}]{nakamoto2008bitcoin}
Nakamoto S (2008) Bitcoin: A peer-to-peer electronic cash system

\bibitem[{Nanda(2017)}]{Perseus}
Nanda V (2017) Perseus: the persistent homology software.
  http://peoplemathsoxacuk/nanda/perseus/indexhtml

\bibitem[{Ober et~al.(2013)Ober, Katzenbeisser, and
  Hamacher}]{ober2013structure}
Ober M, Katzenbeisser S, Hamacher K (2013) Structure and anonymity of the
  bitcoin transaction graph. Future internet 5(2):237--250

\bibitem[{Ron and Shamir(2013)}]{ron2013quantitative}
Ron D, Shamir A (2013) Quantitative analysis of the full bitcoin transaction
  graph. In: Int. Conf. on Financial Cryptography and Data Security, Springer,
  pp 6--24

\bibitem[{Shah and Zhang(2014)}]{shah2014bayesian}
Shah D, Zhang K (2014) Bayesian regression and bitcoin. In: Communication,
  Control, and Computing, 52nd Allerton Conf. on, IEEE, pp 409--414

\bibitem[{Sorgente and Cibils(2014)}]{sorgente2014reaction}
Sorgente M, Cibils C (2014) The reaction of a network: Exploring the
  relationship between the bitcoin network structure and the bitcoin price. No
  Data

\bibitem[{Swanson(2014)}]{swanson2014learning}
Swanson T (2014) Learning from bitcoin's past to improve its future

\bibitem[{Tschorsch and Scheuermann(2016)}]{tschorsch2016bitcoin}
Tschorsch F, Scheuermann B (2016) Bitcoin and beyond: A technical survey on
  decentralized digital currencies. IEEE Comm Surveys 18(3):2084--2123

\bibitem[{Williams and Rasmussen(1996)}]{williams1996gaussian}
Williams CK, Rasmussen CE (1996) Gaussian processes for regression. In: NIPS,
  pp 514--520

\bibitem[{Yang and Kim(2015)}]{yang2015bitcoin}
Yang SY, Kim J (2015) Bitcoin market return and volatility forecasting using
  transaction network flow properties. In: IEEE SSCI, pp 1778--1785

\bibitem[{Zomorodian(2010)}]{Zomorodian:2010}
Zomorodian A (2010) Fast construction of the vietoris-rips complex. Computers
  and Graphics 34(3):263--271

\bibitem[{Zou and Hastie(2005)}]{zou2005regularization}
Zou H, Hastie T (2005) Regularization and variable selection via the elastic
  net. Journal of the Royal Statistical Society: Series B 67(2):301--320

\end{thebibliography}
\end{document}